%% file: neurips_2026.tex
\documentclass{article}

\usepackage{microtype}
\usepackage{graphicx}
\usepackage{subfigure}
\usepackage{booktabs}
\usepackage{hyperref}
\usepackage{amsmath}
\usepackage{amssymb}
\usepackage{mathtools}
\usepackage{amsthm}
\usepackage[hang,flushmargin]{footmisc}
\usepackage{enumitem}
\usepackage{caption}
\usepackage{enumitem}

\usepackage[capitalize,noabbrev]{cleveref}

\usepackage{algorithm}

\usepackage{algpseudocode}
\usepackage{diagbox}

\theoremstyle{plain}
\newtheorem{theorem}{Theorem}[section]
\newtheorem{proposition}[theorem]{Proposition}
\newtheorem{lemma}[theorem]{Lemma}
\newtheorem{corollary}[theorem]{Corollary}
\theoremstyle{definition}
\newtheorem{definition}[theorem]{Definition}
\newtheorem{assumption}[theorem]{Assumption}
\theoremstyle{remark}
\newtheorem{remark}[theorem]{Remark}

 \usepackage[main, final]{neurips_2026}
\usepackage[utf8]{inputenc} % allow utf-8 input
\usepackage[T1]{fontenc}    % use 8-bit T1 fonts
\usepackage{hyperref}       % hyperlinks
\usepackage{url}            % simple URL typesetting
\usepackage{booktabs}       % professional-quality tables
\usepackage{amsfonts}       % blackboard math symbols
\usepackage{nicefrac}       % compact symbols for 1/2, etc.
\usepackage{microtype}      % microtypography
\usepackage{xcolor}         % colors

\title{Mitigating Overgeneralization in RND \\  via Spectral Target Design}

\author{
  \shortstack{
    \textbf{Minseok Jeong}$^{1,*}$ \\
    {\footnotesize\texttt{dsa950115@kaist.ac.kr}}
  }
  \qquad
  \shortstack{
    \textbf{Yechan Lee}$^{1,*}$ \\
    {\footnotesize\texttt{tuuktuc86@kaist.ac.kr}}
  }
  \qquad
  \shortstack{
    \textbf{Hyewon Choi}$^{1}$ \\
    {\footnotesize\texttt{chw110825@kaist.ac.kr}}
  }
  \\[0.5em]
  \shortstack{
    \textbf{Jeongyong Yang}$^{2}$ \\
    {\footnotesize\texttt{seiryu@uw.edu}}
  }
  \qquad
  \shortstack{
    \textbf{SooJean Han}$^{1,\dagger}$ \\
    {\footnotesize\texttt{soojean@kaist.ac.kr}}
  }
  \\[0.5em]
  {\small\normalfont
    $^{1}$KAIST, $^{2}$University of Washington
  }
  \vspace{-0.6em}
}

\begin{document}

\maketitle

\begingroup
\renewcommand{\thefootnote}{}
\footnotetext{
\fontsize{9.0pt}{10pt}\selectfont
$^{*}$Co-first authors. See Appendix \ref{app:contributions} for individual contributions.
\qquad
$^{\dagger}$Corresponding author.
}
\addtocounter{footnote}{-1}
\endgroup

\begin{abstract}
Random Network Distillation (RND) is a scalable novelty signal for RL based on a fixed random target network and a trained predictor. However, it can overgeneralize: the predictor extrapolates the random target off the data manifold, causing novelty scores to collapse on unfamiliar inputs. This paper analyzes this failure through a kernel-theoretic lens. Under a GP target model and an NTK-regime predictor, we show that the RND novelty score is a \textit{cross-kernel residual variance}: it is governed jointly by the \textit{target covariance kernel} and the \textit{predictor NTK}, rather than by either one alone. This identifies overgeneralization as spectral under-excitation. Motivated by this view, we formulate spectral target design as a principle for shaping RND novelty geometry. As a concrete instance, we replace the implicit random-network target with a bandwidth-controlled Random Fourier Feature (RFF) target, redistributing residual signal toward modes less attenuated by the predictor. Experiments on offline D4RL, online Atari, and Fetch manipulation benchmarks show improved novelty discrimination and downstream RL performance. The code for the RL implementation is available at \href{https://github.com/tuuktuc86/RFF-RND}{\underline{\textcolor{blue}{https://github.com/tuuktuc86/RFF-RND}}}.
\end{abstract}

\section{Introduction}

Random Network Distillation (RND)~\citep{burda2018rnd} is a simple yet effective novelty signal for reinforcement learning (RL), using the prediction error of a trained network against a fixed random target to measure familiarity. Its scalability and ease of integration have made it a standard component in modern RL pipelines, driving exploration in online settings~\citep{burda2019large} and enforcing conservatism in offline ones~\citep{rezaeifar2022anti, nikulin2023anti}.

Despite its widespread adoption, RND suffers from a fundamental failure mode known as \emph{overgeneralization}: the predictor can extrapolate beyond the data manifold, assigning low novelty scores to inputs beyond the data coverage radius that should be flagged as novel~\citep{belkin2018overgen}. This undermines the intrinsic reward in both settings --- misallocating exploration bonuses toward already-familiar regions and failing to penalize out-of-support actions sufficiently. Prior work has addressed this through
distributional extensions~\citep{yang2024exploration} or architectural modifications~\citep{nikulin2023anti}, but these approaches do not explicitly characterize \emph{why} overgeneralization occurs or provide principled control over the novelty signal's geometric structure. This raises a natural question:  \[\textit{What governs the discriminative power of RND, and how can it be reliably controlled?}\]
In this work, we answer this question through a kernel-theoretic analysis~\citep{rasmussen2006gaussian,scholkopf2002learning}. We show that overgeneralization arises not from the predictor or target network in isolation, but from their interaction: when the predictor extrapolates the target accurately beyond the data manifold, the novelty score collapses off-manifold. Motivated by this analysis, we propose a spectral target-design principle for RND. 

% Our analysis shows that residual collapse occurs when the target does not sufficiently vary directions that are poorly interpolated by the predictor.
% under-excites modes where predictor residuals would otherwise survive. 

Specifically, directly aligning the target with hard-to-predict modes is impractical, since the predictor spectrum is implicit, architecture-dependent, and generally unknown before training.
We therefore seek predictor-agnostic target families whose spectra can explicitly control the kernel geometry, and are lightweight to sample. 
Random Fourier Features (RFF)~\citep{rahimi2007random, rahimi2008weighted, le2013fastfood} provide one such concrete example, replacing the random MLP target with a target whose spectral density and coordinate-wise independence are directly specified.
% We refer to this instantiation as \texttt{RFF-RND}.
% replace the implicit random MLP target with an explicit RFF target. The resulting method, \texttt{RFF-RND}, requires no additional networks and introduces a single interpretable novelty-scale parameter.
This method provides an explicit control knob for shaping the novelty scale without changing the downstream RL algorithm.

% Through experiments on D4RL offline benchmarks and Atari online benchmarks, we demonstrate that \texttt{RFF-RND} consistently outperforms RND and its variants, with particular gains in sparse-reward and
% out-of-distribution settings where reliable novelty discrimination matters most.

The rest of the paper is organized as follows. 
Section~\ref{sec:problem_setup} defines the coverage-based failure mode we want RND to avoid. 
Section~\ref{sec:theory} analyzes why the residual can collapse by expressing RND energy as a cross-kernel residual variance. Section~\ref{sec:spectral-design} turns this mechanism into a target-design principle, introduces \texttt{RFF-RND}, and provides a frequency-domain interpretation of the same mechanism. Section~\ref{sec:experiment} diagnoses our theory and empirically evaluates our methods on downstream RL tasks.

\section{Problem Setup and Motivation}
\label{sec:problem_setup}

\begin{figure}
    \centering
    \includegraphics[width=0.9\linewidth]{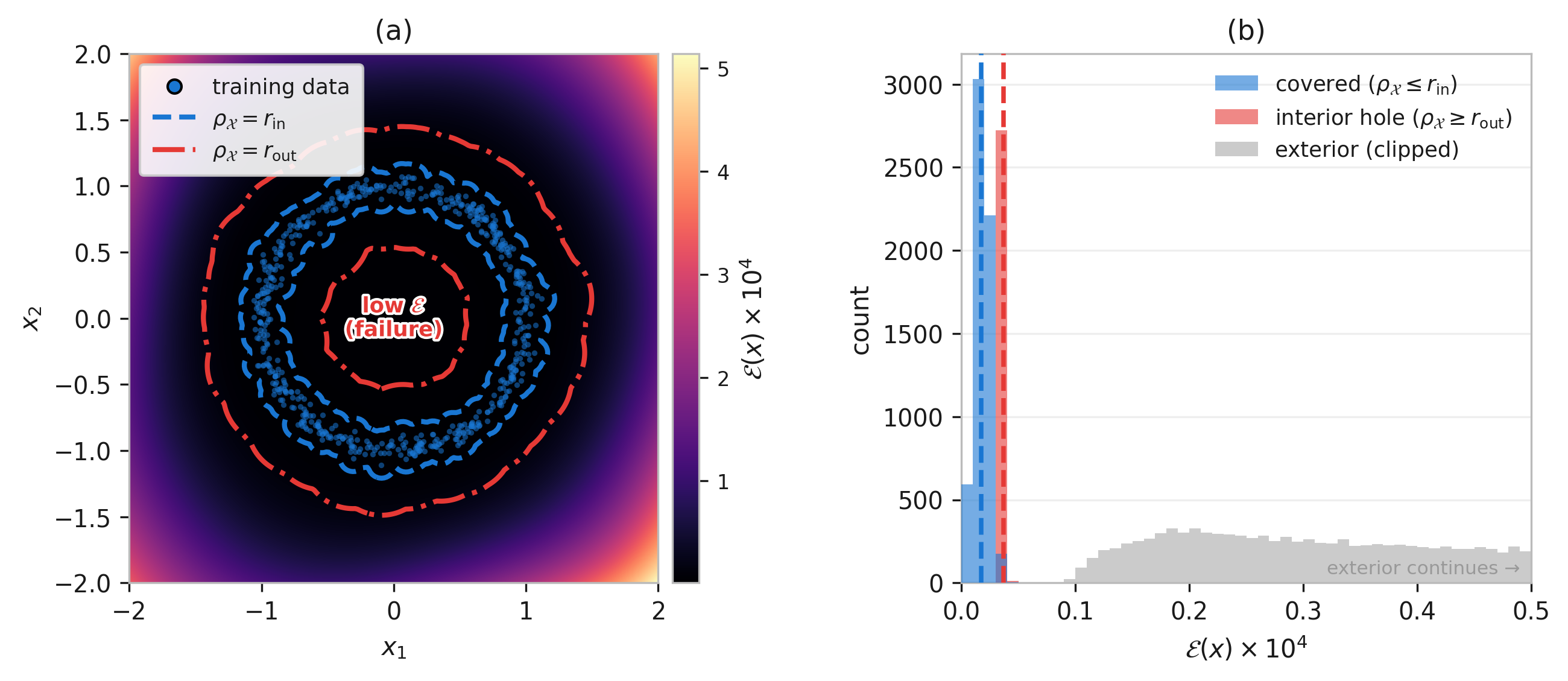}
    \caption{
    \textbf{Coverage-based diagnostic for standard RND on \(\mathbb S^1\) data.} (a) RND energy heatmap with contours defined by \(\rho_{\mathcal D}(x)=\min_{x_i\in\mathcal D}d_{\mathcal X}(x,x_i)\). The blue dashed contour denotes \(\rho_{\mathcal D}=r_{\rm in}\), and the red dash-dotted contour denotes \(\rho_{\mathcal D}=r_{\rm out}\). The interior hole (``low $\mathcal{E}$ (failure)'') lies beyond the coverage radius but receives low RND energy. (b) Energy distributions by region. The covered band (blue, \(\rho_{\mathcal D}\le r_{\rm in}\)) and the interior hole (red, \(\rho_{\mathcal D}\ge r_{\rm out}\)) have overlapping low-energy distributions, illustrating a failure of coverage-based residual separation. The exterior region (gray) is shown only for reference. Energies are scaled by \(10^4\) for visualization since the trained RND energies are small. % do you think you need to explain why they are scaled?
    }
    \label{fig:nearood}
\end{figure}

\subsection{RND for Novelty Detection}\label{subsec:rnd_background} 
Standard RND consists of a fixed target network $f_t:\mathcal{X} \to \mathbb{R}^{m}$ and a trainable predictor network $f_p:\mathcal{X} \to \mathbb{R}^{m}$, both implemented as multi-layer perceptrons (MLPs) with random initialization. Let $x \in \mathcal{X} \subset \mathbb{R}^d$ denote an input datapoint, and $\mathcal D = \{x_i\}_{i=1}^n$ denote the dataset. The predictor $f_p(\cdot;\theta)$ is trained to match the output of the target network by minimizing the squared error on $\mathcal{D}$:
\begin{equation}\label{RND_obj}
    \min_{\theta} \sum_{i=1}^n \mathcal{E}(x_i;\theta),
\end{equation}
where the RND energy $\mathcal{E} : \mathbb{R}^d \to \mathbb{R}_{\ge0}$ is defined as $\mathcal{E}(x_i;\theta) \triangleq \frac{1}{m}\bigl\| f_t(x_i) - f_p(x_i;\theta) \bigr\|_2^2$. When the context is clear, we write $\mathcal E(x) = \mathcal E(x;\theta), f_p(x) = f_p(x ; \theta)$ for brevity, i.e., the RND novelty score at a query point $x$ is $\mathcal E(x) = \frac{1}{m}\| f_t(x) - f_p(x) \|_2^2$.

% DRND~\citep{yang2024exploration} mitigates this failure by replacing a single prediction error with ensemble variance, thereby capturing uncertainty across random target draws. Conditional RND~\citep{nikulin2023anti} instead modulates
% the target via FiLM conditioning, implicitly making the target geometry input-dependent. Both improve discrimination empirically, but neither directly
% controls the target kernel geometry: DRND inherits the shared architectural bias of its ensemble members, while Conditional RND leaves the locality scale as an
% implicit byproduct of the conditioning architecture.

Although standard RND assigns high energy to some exterior regions, it assigns low energy to the region that is poorly covered by the observed data. Thus, reliable novelty detection requires a coverage-based criterion rather than merely assigning high energy somewhere off the data manifold.

\subsection{A Coverage-Based Separation Criterion}
\label{subsec:desideratum}

% Novelty detection does not require large residuals everywhere. 
A useful RND score should separate well-covered inputs from poorly-covered ones, i.e., the target should be predictable near visited samples but difficult to extrapolate beyond the data coverage radius. Let \(d_{\mathcal X}\) be a task-relevant metric or representation-induced distance on \(\mathcal X\).
We formalize our desired novelty behavior as follows. 

\begin{definition}[Coverage-based residual separation]
\label{def:coverage_residual_separation}
Let $\rho_{\mathcal D}(x) \triangleq \min_{x_i\in \mathcal D} d_{\mathcal{X}}(x, x_i)$. We say that RND satisfies \emph{coverage-based residual separation} if there exist \(0\le r_{\rm in}<r_{\rm out}\) and \(0\le\epsilon<\gamma\) such that
\[
\rho_{\mathcal D}(x) \leq r_{\rm in} \;\Rightarrow\; \mathcal{E}(x) \leq \epsilon,
\qquad
\rho_{\mathcal D}(x) \geq r_{\rm out} \;\Rightarrow\; \mathcal{E}(x) \geq \gamma.
\]
\end{definition}
% This definition serves as a conceptual target; our theoretical
% results in Section~\ref{sec:spectral-design} establish a spectral
% analogue in an idealized setting.

We use this criterion as a diagnostic notion rather than a universal guarantee: both
the metric \(d_{\mathcal X}\) and the relevant coverage parameters $r_{\rm{in}}, r_{\rm{out}}, \epsilon, \gamma$ are task-dependent. In Section~\ref{sec:spectral-design}, we establish a spectral analogue in an idealized setting.
% the target should be predictable from nearby visited states, while becoming difficult to interpolate once the query point moves beyond the coverage radius of the data. 
Figure~\ref{fig:nearood} intuitively illustrates this coverage-based failure mode on an $\mathbb{S}^1$ toy dataset.

\subsection{RND Variants through the Lens of Kernel Geometry}

We next situate this coverage-based failure mode relative to existing RND variants.
% Despite this desired coverage-based behavior, standard RND can suffer from overgeneralization.

\paragraph{Predictor- and target-focused RND variants.}
Many RND variants implicitly modify one side of the kernel geometry. Conditional RND~\citep{nikulin2023anti} conditions the predictor via FiLM-style modulation~\citep{devries2017modulating, perez2018film, dumoulin2018featurewise}, inducing a condition-dependent feature map that deforms the resulting geometry. In contrast, Distributional RND (DRND)~\citep{yang2024exploration} acts on the target side. It replaces the prediction error induced by a single target network with the ensemble variance induced by multiple target networks, thereby capturing uncertainty across random target draws. Although these approaches showed favorable performance improvements, they did not identify the underlying cause.

\paragraph{Why cross-kernel geometry matters.}
These variants suggest that \emph{both} sides of the RND residual matter: the predictor geometry determines how target values are interpolated, while the target geometry determines what variation is available to be detected. In the following sections, we make this interaction explicit by characterizing RND energy as a cross-kernel residual variance.

\section{RND Energy as a cross-kernel residual variance} % Theoretical Analysis
\label{sec:theory}
% \paragraph{Analytical model.}
We analyze RND under two idealizations: a GP model for the random target and an NTK approximation for the predictor. 

% These assumptions are not meant to exactly describe finite networks, but to isolate the two geometric objects that govern the residual.

% : the target kernel and the predictor interpolation kernel.

\begin{assumption}[GP target model]
\label{asm:gp_target}
We model each output coordinate of $f_t$ as an independent GP:
\[
f_t^{(j)} \sim \mathcal{GP}(0, k_t), \quad j=1,\dots,m,
\]
with output coordinates mutually independent.    
\end{assumption}
Assumption \ref{asm:gp_target} is a natural infinite-width idealization of standard RND: a randomly initialized MLP converges to a GP with the associated NNGP kernel~\citep{neal1996bayesian,lee2018deep, matthews2018gpbehaviour}. A key consequence of this is \emph{joint Gaussianity}, that for any query $x$ and dataset inputs $X$, the collection $\{f_t(x), f_t(x_1), \dots, f_t(x_n)\}$ is jointly Gaussian with cross-covariances 
\[\mathbb E[f_t(x) f_t(x_i)^\top] = k_t(x, x_i) \, I_m.\] Equivalently, for any inputs $x, x' \in \mathcal X$, $\mathbb E[f_t(x) f_t(x')^\top] = k_t(x,x') \, I_m$. The induced covariance on the dataset is $K_t = k_t(X,X) \in \mathbb R^{n \times n}$.

\begin{assumption}[NTK predictor model]
\label{asm:ntk_predictor}
The predictor is analyzed in the lazy-training regime
~\citep{chizat2019lazy,lee2019wide}. We use a decoupled multi-output NTK
approximation, where the squared loss separates across the \(m\) output
coordinates and each coordinate is governed by the same scalar NTK
\(\Theta_p\)~\citep{jacot2018neural,du2019gradient,allen2019convergence,
arora2019exact}. 
% Equivalently, the vector-valued predictor is obtained by
% applying the corresponding scalar kernel regression operator column-wise to the target matrix.
\footnote{A fully vector-valued NTK would replace \(\Theta_p\) with an \(nm \times nm\) block kernel over input--output coordinate pairs. We use the decoupled form only to avoid unnecessary notation: the same interpolation argument applies coordinate-wise, while the analysis below focuses on the induced dependence on the target matrix \(F_X\).} Under gradient flow with \(\ell_2\) regularization, this yields kernel ridge regression with respect to \(\Theta_p\):
\[
\Theta_p = \Theta_p(X,X) \in \mathbb R^{n \times n},
\quad
\Theta_p(x,X) \in \mathbb R^{1 \times n}.
\]
Define the target matrix
% $F_X = [f_t(x_1)^\top , \dots, f_t(x_n)^\top] \in \mathbb R^{n \times m}$. 
$F_X = [f_t(x_1), \dots, f_t(x_n)]^\top \in \mathbb R^{n \times m}$. 
The limiting predictor at a query point \(x\) is $f_p^\infty(x)
\triangleq
F_X^\top(\Theta_p+\lambda I)^{-1}\Theta_p(x,X)^\top \in \mathbb{R}^{m}$.\footnote{
Standard RND training does not include explicit ridge regularization. Here,
\(\lambda>0\) is introduced to ensure that \(\Theta_p+\lambda I\) is invertible;
the limit \(\lambda\to0^+\) recovers the minimum-norm interpolant. We keep
\(\lambda>0\) throughout for analytical clarity.
} For brevity, define
\[
A_x \triangleq \Theta_p(x,X)(\Theta_p + \lambda I)^{-1} \in \mathbb R^{1 \times n},
\]
so that $f_p^\infty(x) = F_X^\top A_x^\top$. 
\end{assumption}

% \begin{remark}[Coordinate-wise independence]
% The form \(f_p^\infty(x)=F_X^\top A_x^\top\) implies coordinate-wise decoupling: $f_p^{\infty,(j)}(x)=F_X^\top A_x^\top^{(j)}, r^{(j)}(x)=f_t^{(j)}(x)-F_X^\top A_x^\top^{(j)}$, where \(F_X^{(j)}\) is the \(j\)-th column of \(F_X\). Because \(A_x\) is
% deterministic given \(X\) and \(\Theta_p\) and shared across \(j\), the i.i.d. GP target coordinates \(\{f_t^{(j)}\}_{j=1}^m\) yield independent residual coordinates \(\{r^{(j)}(x)\}_{j=1}^m\).
% \end{remark}

Although Assumption~\ref{asm:ntk_predictor} is idealized, it provides a natural approximation for RND.
% : after the random target is fixed, predictor training reduces to supervised regression from \(x_i\) to \(f_t(x_i)\).
Because the predictor fits a fixed random target, the supervision signal contains no coherent task-level structure to be extracted through representation learning. So, that predictor only requires adapting the target kernel. A similar perspective has been studied in recent work connecting RND to ensemble and Bayesian interpretations in the infinite-width regime~\citep{zanger2026equivalencerandomnetworkdistillation}. However, our focus is complementary: we use the NTK view to diagnose why RND can fail as a novelty estimator.  We begin by defining the cross-kernel residual variance.
\begin{definition}[cross-kernel residual variance]
\label{def:cross_kernel}
Given target kernel $k_t$ and predictor NTK $\Theta_p$, the \emph{cross-kernel residual variance} at $x$ is
\[
\sigma^2_{\mathrm{cross}}(x;\, k_t, \Theta_p)
\;\triangleq\;
k_t(x,x) - 2A_x\, k_t(X,x) + A_x K_t A_x^\top.
\]
\end{definition}
We next define the finite-output RND energy under the NTK predictor, which surrogates $\mathcal{E}(x)$, by 
\[
\mathcal{E}_m(x) \triangleq\frac{1}{m}\|f_t(x)-f_p^\infty(x)\|_2^2.
\] 
This leads to the following characterization of its infinite-output limit.

% \footnote{The output dimension \(m\) controls concentration of the RND energy
% around its kernel-level limit; see Appendix~\ref{app:finite_dim_verification}.}

\begin{theorem}[Infinite-dimensional GP target RND energy]
\label{thm:main}
Under the GP target model and NTK-regime predictor,
\[
\mathcal E_m(x)
\xrightarrow[m\to\infty]{a.s.}
\sigma^2_{\mathrm{cross}}(x;k_t,\Theta_p).
\]
% where we define the ideal GP/NTK-regime RND energy $\mathcal{E}_{\infty}(x) \triangleq \lim_{m\to \infty}\mathcal{E}_m(x)$.

\end{theorem}

The proof is provided in Appendix~\ref{app:proof_main}. Note that, although our proof exploits the Gaussianity of the target GP, this assumption is stronger than necessary. In fact, the energy formula depends only on the second-order covariance structure \(k_t\) of the target process. In Section~\ref{sec:spectral-design}, we show that RFF targets implement the same
covariance structure in expectation, despite not being GP sample paths.

Theorem~\ref{thm:main} is the key point of our analysis. It shows that,  in the
infinite-output GP/NTK limit, the RND energy is cross-kernel residual variance, i.e., $\mathcal{E}_\infty(x) = \sigma^2_{\mathrm{cross}}(x;k_t,\Theta_p)$. The target kernel \(k_t\) determines the covariance of the random target values, whereas the predictor NTK \(\Theta_p\) determines the interpolation weights \(A_x\). These weights specify how the predictor combines the observed target values \(F_X\) to estimate the target at a query point \(x\). Hence the RND
score is governed by the interaction between target variation and predictor interpolation, rather than by either kernel alone.

\section{Spectral Target Design}
\label{sec:spectral-design}
% Section~\ref{sec:theory} shows that RND energy is a cross-kernel residual variance, with the target covariance and the predictor NTK. 
We now use this view to derive a target-design principle. The key idea is to treat the target covariance as a spectral design object: rather than inheriting an uncontrolled spectrum from a random network, we seek a target family with an explicit knob for shaping RND energy landscapes.

\subsection{Residual Collapse as Spectral Under-Excitation}

% We next use it to design target kernels whose variation is less easily erased by predictor interpolation, by analyzing
We analyze the induced training residual $R_X = F_X-\widehat{F}_X, \widehat{F}_X=[f_p^\infty(x_1), \dots , f_p^\infty(x_n)]^\top$ whose $\frac{1}{m}\operatorname{tr}(R_X R_X^\top)$ is the sum of RND energies over the training set.

\begin{lemma}[Spectral residual decomposition]
\label{cor:spectral}
Consider the spectral decomposition $\Theta_p = U_p \Lambda_p U_p^\top$ with $\Lambda_p{\,=\,}\mathrm{diag}(\lambda_{p,1}, \dots, \lambda_{p,n})$ and let $u_i$ denote the $i$-th column of $U_p$. The training residual satisfies $R_X{\,=\,}\lambda(\Theta_p + \lambda I)^{-1} F_X$, and $\mathbb{E}\|u_i^\top R_X\|_2^2 = m \!\left(\frac{\lambda}{\lambda_{p,i}+\lambda}\right)^{\!2} u_i^\top K_t\, u_i$.
\end{lemma}

% \begin{proof}
% Since $R_X = [I - \Theta_p(\Theta_p+\lambda I)^{-1}]F_X
% = \lambda(\Theta_p+\lambda I)^{-1}F_X \triangleq B_\lambda F_X$,
% and $\mathbb{E}[F_X F_X^\top] = m \cdot K_t$,
% \[
% \mathbb{E}[R_X R_X^\top]
% = m \cdot B_\lambda K_t B_\lambda
% = m \cdot U_p D_\lambda U_p^\top K_t U_p D_\lambda U_p^\top,
% \]
% where $D_\lambda = \mathrm{diag}(\lambda/(\lambda_{p,i}+\lambda))$.
% Projecting onto $u_i$ gives the result.
% % \end{proof}

A detailed proof is provided in Appendix~\ref{app:proof_cor_spectral}. Lemma~\ref{cor:spectral} shows that the regularized residual on the training set decomposes naturally in the eigenbasis of the predictor NTK $\Theta_p$. In addition, since $K_t$ governs the target covariance globally, the same spectral structure determines residual behavior at query points beyond the training set.
% Corollary~\ref{cor:spectral} shows that the RND residual energy 
% It identifies which target modes survive predictor interpolation and thus provides a spectral diagnostic for residual collapse.
% The RND residual energy $\mathcal{E}_m(x)$ then decomposes mode by mode into a predictor-side filter and a target-side variance term.
% It identifies which target modes survive predictor interpolation and thus provides a spectral diagnostic for residual collapse. 
We can further decompose the training-set residual energy along each predictor NTK eigenmode into a predictor-side and a target-side term.
% A detailed proof is in Appendix~\ref{app:proof_cor_spectral}. 
% Lemma~\ref{cor:spectral} decomposes the training-set residual 
% in the eigenbasis of the predictor NTK, identifying which target modes 
% survive predictor interpolation. Although stated for the training-set residual, the same spectral structure of $K_t$ governs how the target varies at query points: 
% targets with low power in predictor-hard modes fail to produce 
% residual signal both on- and off-manifold.
\[
\boxed{\text{Predictor's mode-}u_i\text{ contribution to RND energy}{
\;\propto\;
\underbrace{\left(\frac{\lambda}{\lambda_{p,i}+\lambda}\right)^2}_{\text{predictor residual filter}}
\;\times\;
\underbrace{u_i^\top K_t u_i}_{\text{target variance in mode }u_i}}}
\]

% \section{Spectral Target Design}
% \label{sec:spectral-design}
% Section~\ref{sec:theory} shows that RND energy is a cross-kernel residual variance, with the target covariance determining what variation is present and the predictor NTK determining how this variation is interpolated. We now use this view to derive a target-design principle. The key idea is to treat the target covariance as a spectral design object: rather than inheriting an uncontrolled spectrum from a random network, we seek a target family with an explicit knob for shaping RND energy landscapes.
% \subsection{Residual Collapse as Spectral Under-Excitation}

If \(u_i\) is a predictor-easy mode \((\lambda_{p,i}\gg \lambda)\), the residual filter nearly vanishes, so even a large value of \(u_i^\top K_t u_i\) contributes little to the residual. Conversely, for predictor-hard modes
\((\lambda_{p,i}\ll \lambda)\), the filter remains large; however, when \(u_i^\top K_t u_i\) is small, the target contains little variation along that mode, leaving little residual signal. In this terminology, overgeneralization arises from predictor-hard modes with insufficient target variation.
% when the target has little power in modes that survive the predictor residual filter.
%
% In the infinite-width view, the target covariance is governed by the NNGP kernel, while predictor interpolation is governed by the NTK. 
% Although target's NNGP and predictor's NTK kernels are not identical, they can share smoothness and global similarity biases. 
 % the target Gram matrix may concentrate its power on modes that are easy for the predictor to interpolate, while placing little power on modes where residuals would survive. 
This motivates framing target design as a mode-wise spectral allocation problem, which we formalize in the following definition.

\begin{definition}[Target power and residual contribution]
\label{def:target_power_and_res_cont}
Let \(\Theta_p = U_p\Lambda_pU_p^\top\), with eigenvectors
\(\{u_i\}_{i=1}^n\) and eigenvalues $\{\lambda_{p,i}\}_{i=1}^n$. For a target Gram matrix \(K_t\), define the \emph{target power}
assigned to predictor mode \(u_i\) by
% \[
$q_i(K_t) \triangleq u_i^\top K_t u_i$.
% \]
For ridge parameter \(\lambda>0\), define the \emph{residual contribution} of mode
\(u_i\) by $r_i(K_t) \triangleq \left(\frac{\lambda}{\lambda_{p,i}+\lambda}\right)^2 q_i(K_t)$.
\end{definition}

In standard RND, the target and predictor networks typically share the same architecture, with their parameters independently initialized from the same distribution. Consequently, their associated kernels are expected to exhibit similar spectral densities.

\subsection{RFF-RND: Target Design as Spectral Allocation}
So far, we have analyzed how the RND energy, particularly its spectral properties, is characterized and determined by the interplay between the predictor and target networks. However, the preceding decomposition suggests an ideal but impractical allocation rule: choose \(K_t\) so that \(q_i(K_t)\) is not concentrated only on predictor-easy modes, but also assigns power to modes where the residual filter is non-negligible. In general, directly optimizing \(K_t\) in the predictor eigenbasis is not practical in standard RND, since \(\Theta_p\), its eigenvectors, and its spectrum are implicit and architecture-dependent. Therefore, a natural extension is to make the target structure explicit in a \emph{predictor-agnostic} form, with a tunable spectrum for generality rather than inheriting an uncontrolled spectrum from a randomly initialized MLP.

Moreover, the preceding analysis only depends on the \textit{target covariance structure}.
% , not on GP sample paths themselves.
This motivates replacing the target family with a shift-invariant \emph{RFF target} (which we call \texttt{RFF-RND}), whose spectral density can be controlled directly.

\begin{corollary}[Infinite-output RFF target RND energy]
\label{cor:rff-energy}
For the $m$-dimensional RFF target $f_t^{(j)}(x)=\sqrt{2}\cos(\omega_j^\top x+b_j)$, $\omega_j\sim p_\alpha$\footnote{Here, we assume $p_\alpha$ is symmetric, i.e., $p_\alpha(\omega)=p_\alpha(-\omega)$. For the RBF target kernel, \(p_\alpha(\omega)=\mathcal N(0,\alpha^{-2}I_d)\) and \(k_t^\alpha(x-x')=\exp(-\|x-x'\|_2^2/(2\alpha^2))\). Other variants of shift-invariant kernels are summarized in Appendix~\ref{app:shift_invariant_kernels}.}, $b_j\sim \mathrm{Unif}[0,2\pi]$, with independent draws across \(j \in \{1,\dots, m\}\), the target coordinates are zero mean and satisfy $\mathbb E_{\omega,b}\!\left[
f_t^{(j)}(x)f_t^{(j)}(x')\right]=k_t^\alpha(x-x')$. Therefore, under the same NTK-regime predictor, $\mathcal E_m(x) \xrightarrow[m\to\infty]{a.s.} \sigma^2_{\mathrm{cross}}(x;k_t^\alpha,\Theta_p)$.
\end{corollary}

% This motivates replacing the target with a shift-invariant \emph{RFF target} (which we call \texttt{RFF-RND}), whose spectral density can be controlled directly. For
% \(j=1,\ldots,m\), let
% \(f_t^{(j)}(x)\triangleq\sqrt{2}\cos(\omega_j^\top x+b_j)\), with
% \(\omega_j\sim p_\alpha\) and \(b_j\sim{\rm Unif}[0,2\pi]\). Then
% \[
% \mathbb E_{\omega,b}\!\left[f_t(x)f_t(x')^\top\right]
% =
% k_t^\alpha(x-x')I_m,\qquad
% k_t^\alpha(\tau)\triangleq\int e^{i\omega^\top\tau}p_\alpha(\omega)d\omega.
% \]
% Thus, the frequency law \(p_\alpha\) directly specifies the target kernel\footnote{Here, we assume $p_\alpha$ is symmetric, i.e., $p_\alpha(\omega)=p_\alpha(-\omega)$, so that the induced shift-invariant kernel \(k_t^\alpha(\tau)\) is real-valued. For the RBF target kernel, \(p_\alpha(\omega)=\mathcal N(0,\alpha^{-2}I_d)\) and \(k_t^\alpha(x-x')=\exp(-\|x-x'\|^2/(2\alpha^2))\). Other variants of shift-invariant kernels are summarized in Appendix~\ref{app:shift_invariant_kernels}.}. We then empirically measure target power in the eigenbasis of the empirical predictor NTK:
% \[
% q_i(K_t^\alpha)=u_i^\top K_t^\alpha u_i,
% \qquad
% (K_t^\alpha)_{ij}=k_t^\alpha(x_i,x_j).
% \]

\begin{figure}
    \centering
    \includegraphics[width=1.0\linewidth]{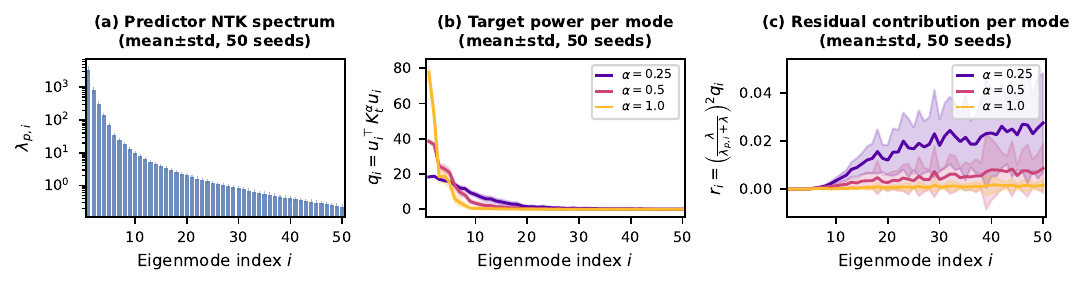}
    \caption{
    \textbf{Spectral allocation of RBF-RFF targets across predictor NTK modes.} (a) Predictor NTK eigenvalues \(\lambda_{p,i}\), sorted in descending order. (b) Target power \(q_i(K_t^\alpha)=u_i^\top K_t^\alpha u_i\) along predictor NTK
    eigenmodes. (c) Mode-wise residual contribution
    \(r_i(K_t^\alpha)=\left(\lambda/(\lambda_{p,i}+\lambda)\right)^2
    q_i(K_t^\alpha)\).
    }
    \label{fig:spectral-allocation}
\end{figure}

The proof is provided in Appendix \ref{app:proof_cor:rff-energy}. As briefly mentioned earlier, here we exploit the fact that it suffices to match only the second-order moments encoded by the target network's Gram matrix, without requiring Gaussianity of the target sample paths. For an RFF target, the corresponding covariance is \(K_t^\alpha\), whose bandwidth \(\alpha\) controls the spectral spread of the target kernel. 

Therefore, we can easily compare target designs by examining how \(K_t^\alpha\) allocates power across the empirical predictor NTK eigenbasis, yielding a substantially more flexible target design.

Figure~\ref{fig:spectral-allocation} illustrates this behavior empirically. Consider an RND setup with a large-width MLP predictor and an RFF target. As expected, the predictor NTK spectrum is concentrated on leading modes, so the residual filter strongly suppresses these modes while leaving higher-index modes less attenuated. The RFF bandwidth controls how target power, and consequently residual contribution, is allocated across this basis. Larger bandwidths concentrate power on easy leading modes ($\alpha=1$), whereas smaller bandwidths spread power toward less-attenuated higher-index modes ($\alpha=0.25$). Specifically, Figure~\ref{fig:spectral-allocation}-(c) shows that smaller RFF bandwidths concentrate residual contribution on high-index NTK modes, where small predictor eigenvalues imply weak interpolation.

% producing larger filtered residual contributions (Figure \ref{fig:spectral-allocation}-(c)). Thus, RFF targets provide an explicit spectral knob for controlling which predictor modes contribute to the residual signal.

\begin{remark}[Pretrained feature map]
Pretrained features can also define a target kernel $K_{\rm pre}(i,j)=\phi_{\rm pre}(x_i)^\top\phi_{\rm pre}(x_j)$. However, the mode-wise power \(u_i^\top K_{\rm pre}u_i\) is inherited from pretraining and is not directly tunable for novelty detection. This is useful
when the pretrained invariances match the desired coverage, but unlike RFF
targets, it does not provide an explicit novelty-scale parameter.
\end{remark}

\subsection{Frequency-Domain Interpretation of Residual Separation}\label{sec:frequency-domain}

% Introducing RFF targets gives us an explicit target spectrum, but the resulting
% RND residual is not determined by this spectrum alone. It depends on how the
% target spectrum is filtered by the predictor interpolation geometry. To make
% The empirical NTK decomposition above applies directly to finite datasets, but it does not yield a closed-form frequency expression. Here, we interpret the same mechanism in a frequency-domain setting,
% this interaction explicit, we c
% The finite-sample NTK decomposition above is exact for arbitrary datasets, but its modes are the empirical eigenvectors of \(\Theta_p\), which need not have an immediate frequency interpretation. To make the spectral mechanism explicit, we consider a clean idealized setting where this eigenbasis is known. On a uniform periodic grid, shift-invariant predictor and target kernels induce circulant
% Gram matrices, and hence both are diagonalized by the Fourier basis. In this case, the general predictor-mode decomposition reduces to a frequency-wise decomposition.
The finite-sample NTK decomposition above is general, but its modes are the empirical eigenvectors of \(\Theta_p\), which need not have an immediate frequency interpretation. To make the spectral mechanism explicit, we consider a clean idealized setting where this eigenbasis is known. On a uniform periodic grid, shift-invariant predictor and target kernels are diagonalized by the Fourier basis. In this case, the general predictor-mode decomposition reduces to a frequency-wise decomposition.

% This reveals how the RFF spectral density \(p_\alpha\) interacts with the predictor spectrum.

\begin{proposition}[Frequency-wise residual decomposition]
\label{prop:frequency-residual}
On a uniform periodic grid, suppose \(\Theta_p\) and \(k_t^\alpha\) are
shift-invariant, with Fourier eigenvalues
\(\widehat{\theta}_p(\omega)\) and \(\widehat{k}_t^\alpha(\omega)\). Then
\[
r_\alpha(\omega)
\triangleq
\frac{1}{m}\mathbb E\|u_\omega^* R_X\|_2^2
=
\left(
\frac{\lambda}{\widehat{\theta}_p(\omega)+\lambda}
\right)^2
\widehat{k}_t^\alpha(\omega).
\]
\end{proposition}

% The proof is given in Appendix~\ref{app:proof_frequency_residual}.
% Proposition~\ref{prop:frequency-residual} gives the frequency-domain counterpart
% of the residual decomposition in Lemma~\ref{cor:spectral}. The residual
% at frequency \(\omega\) factors into a predictor-side spectral residual filter, $(\lambda / (\widehat{\Theta}_p(\omega)+\lambda))^2$, and a target-side spectral power term, \(\widehat{k}_t^\alpha(\omega)\). The filter is small where the predictor kernel has large spectral mass and larger at frequencies that the predictor interpolates weakly. Thus, a frequency contributes strongly to the RND residual only when the target assigns power to that frequency and the predictor does not interpolate it away. In this idealized shift-invariant setting, this gives a spectral perspective on the coverage-based criterion of Definition~\ref{def:coverage_residual_separation}.

The proof is given in Appendix~\ref{app:proof_frequency_residual}. Proposition~\ref{prop:frequency-residual} is the frequency-domain counterpart of the residual decomposition in Lemma~\ref{cor:spectral}. It shows that the residual at frequency \(\omega\) factors into a predictor-side spectral residual filter, \((\lambda/(\widehat{\theta}_p(\omega)+\lambda))^2\), and a target-side spectral power term, \(\widehat{k}_t^\alpha(\omega)\). This suggests that target should allocate power \emph{broadly enough} to cover frequencies that are weakly interpolated by the predictor, rather than concentrating power on a
narrow, easily interpolated band.

\begin{figure*}[t]
    \centering
    \includegraphics[width=1.0\linewidth]{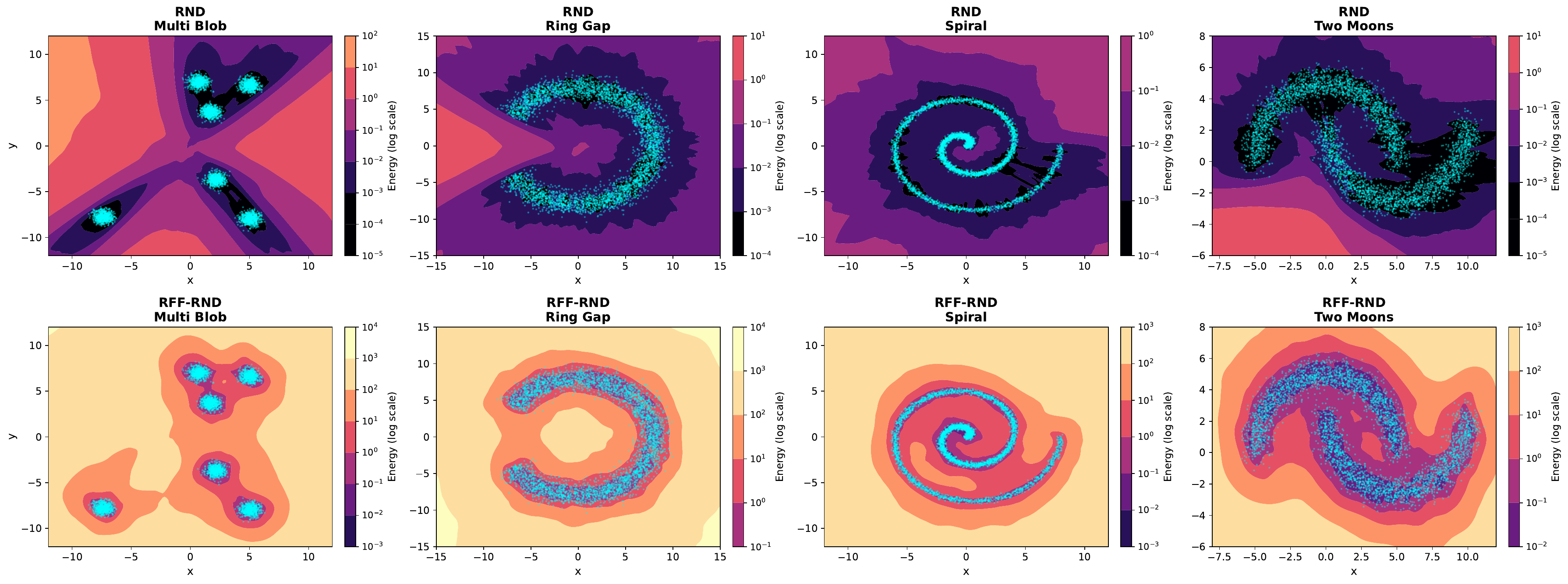}
    % \captionsetup{skip=-1pt}
    \caption{\textbf{Energy landscapes (\(\alpha=0.75\), \(m=64\)).} Top: standard RND exhibits overly confident off-manifold energy patterns. Bottom: \texttt{RFF-RND} produces more consistent landscapes with clearer separation between the data manifold and off-manifold regions.}
    \label{fig:rnd_energy_heatmap_4manifolds}
    % \vspace{-2.0em}
\end{figure*}

The same mechanism can also be understood intuitively in the spatial domain. For RBF-RFF targets, smaller \(\alpha\) makes \(k_t^\alpha(x,x')=\exp(-\|x-x'\|_2^2/(2\alpha)^2)\) decay more rapidly with distance, so target values far from \(\mathcal D\) are harder to infer from observed dataset values. More explicitly, Corollary~\ref{cor:spectral} gives 
\[\mathcal{E}_\infty(x) = \sigma^2_{\mathrm{cross}}(x;\,k_t^\alpha,\Theta_p) = 1 - 2A_x k_t^\alpha(X,x) + A_x K_t^\alpha A_x^\top.\]
For out-of-coverage points, the cross-covariance term \(A_x k_t^\alpha(X,x)\) is driven toward zero, while the quadratic term \(A_x K_t^\alpha A_x^\top\) can remain non-negligible even when \(A_x\) is small. The resulting residual increase induces the coverage-based separation in Definition~\ref{def:coverage_residual_separation}, consistent with the empirical diagnostic in Figure~\ref{fig:rnd_energy_heatmap_4manifolds}.

\section{Experiments} \label{sec:experiment}

\subsection{Diagnostics and Ablations for \texttt{RFF-RND}}
\label{subsec:diagnostic_and_ablations_for_rffrnd}
We provide diagnostics and ablations for \texttt{RFF-RND}, validating the kernel-level predictions and examining the effects of \(\alpha\), \(m\), and predictor network width \(h\) on the induced novelty geometry.
To keep the main text focused on the empirical narrative, we summarize these diagnostic results here and defer the full ablations to the appendices.

\paragraph{Finite-output concentration.}
Appendix~\ref{app:ablation_rffrnd} verifies the finite-output concentration predicted by Corollary~\ref{cor:rff-energy}. 
% This bridges the \(m\to\infty\) theory with finite-dimensional RND in practice. In summary, 
We observed that as the RFF output dimension \(m\) increases, the empirical energy \(\mathcal E_m(x)\) aligns more closely with the kernel-level prediction \(\sigma^2_{\mathrm{cross}}(x;k_t^\alpha,\Theta_p)\).
On the Two Moons diagnostic, Pearson agreement increases from about \(r=0.65\) at
\(m=4\) to about \(r=0.97\) at \(m=64\), suggesting that moderate output dimensions already provide reliable kernel-level estimates.

% \paragraph{Finite-predictor robustness.}
% \textcolor{red}{The NTK analysis idealizes the predictor as infinitely wide, whereas our
% implemented predictors are finite MLPs.
% Appendix~C.3 varies the predictor width over
% $h\in\{16,32,64\}$.
% Across this range, width primarily changes contour smoothness, while the
% bandwidth-dependent novelty scale and manifold-aware separation remain
% qualitatively stable.
% Together with the finite-$m$ concentration result above, this suggests that
% the predicted target--predictor geometry remains informative at the finite
% dimensions used in practice, although it does not constitute a general
% feature-learning guarantee.}

\paragraph{Frequency-domain diagnostic.}
Appendix~\ref{app:frequency_residual_verification} visualizes the frequency-wise
decomposition in Proposition~\ref{prop:frequency-residual}. This diagnostic directly checks the product structure of the residual, rather than downstream performance.
% To summarize the results, 
We observe the RFF bandwidth controls the target spectrum, while the predictor
NTK induces a residual filter that suppresses frequencies it interpolates well. The observed residual power appears where target spectral power overlaps with frequencies that pass this filter, supporting bandwidth as a spectral knob for shaping the RND residual.

\paragraph{Energy-landscape ablations.}
Appendix~\ref{app:ablation_energy} studies how the RFF bandwidth
\(\alpha\), output dimension \(m\), and predictor width \(h\) affect the energy
landscape. The bandwidth \(\alpha\) controls the spatial scale of novelty: small values produce localized but noisy landscapes, while large values over-smooth the energy and blur nearby manifold boundaries. Increasing \(m\) stabilizes the finite-output energy, whereas varying \(h\) mainly changes the smoothness of the fitted contours without altering the overall manifold-aware separation. Together, these ablations clarify the roles of the main design parameters in
our diagnostics: \(\alpha\) changes the spatial scale of the energy landscape,
\(m\) improves finite-output stability, and \(h\) mainly affects the smoothness
of the fitted contours over the tested range.

% changing $m$ determines the discrepancy of energy v.s. true cross kernel variance $\sigma^2_{\rm{cross}}$. The output dimension \(m\) controls concentration of the RND energy around its kernel-level limit;

% \paragraph{Comparison with other RND variants.}
% Appendix~\ref{app:overconf_MLP_family} provides additional synthetic diagnostics for DRND and input-conditioned RND variants. DRND improves contrast through a distributional correction, while conditioned RND reshapes the energy landscape through gating, bilinear, or FiLM-style modulation. However, these mechanisms leave the induced novelty geometry implicit and architecture-dependent, in contrast to \texttt{RFF-RND}'s explicit target-kernel control.

\subsection{Additional Comparisons with RND Variants}
% Appendix~\ref{app:overconf_MLP_family} compares \texttt{RFF-RND} with DRND
% and input-conditioned RND variants on synthetic manifolds, including
% computational and memory costs. DRND improves novelty contrast through a distributional correction over $N$
% frozen target networks, while input-conditioned RND reshapes the energy
% landscape through gating, bilinear, or FiLM-style modulation.
% However, their novelty geometry remains implicit and architecture-dependent:
% DRND's kernel is determined by the MLP function class at initialization,
% while input-conditioned variants depend on the choice of modulation
% architecture. These designs also introduce additional overhead---DRND stores and evaluates
% all $N$ target networks at inference, while input-conditioned RND adds
% predictor-side modulation modules. In contrast, \texttt{RFF-RND} replaces the random target network with fixed
% RFF parameters $\Omega\in\mathbb{R}^{d\times m}$ and $b\in\mathbb{R}^m$, providing explicit target-kernel control with comparatively small target-side overhead.

Appendix~\ref{app:overconf_MLP_family} compares \texttt{RFF-RND} with DRND
and input-conditioned RND variants on synthetic manifolds, including computational and memory costs. DRND improves novelty contrast through a distributional correction over $N$ frozen target networks, while input-conditioned RND reshapes the energy landscape through gating, bilinear, or FiLM-style modulation. However, their novelty geometry remains implicit and architecture-dependent, and both introduce additional overhead. In contrast, \texttt{RFF-RND} replaces the random target network with fixed RFF parameters $\Omega\in\mathbb{R}^{d\times m}$ and $\mathbf b\in\mathbb{R}^m$, providing explicit target-kernel control with comparatively small target-side overhead.

\subsection{Downstream Performance  on RL}
% In this subsection, 
% we present ablations of \texttt{RFF-RND} and evaluate it in both online and offline RL.
% we conduct experiments and ablation studies to empirically validate our earlier theoretical insights on the properties of novelty discrimination in our proposed approach, especially \texttt{RFF-RND}. 
We compare our method across a range of online and offline RL benchmarks to assess its effect on downstream task performance.
%
% \paragraph{MDP and intrinsic signal.}
We consider discounted MDPs~\citep{sutton2018reinforcement} \(\mathcal M{\,=\,}(\mathcal S,\mathcal A,P,r,\gamma)\) with objective \(J(\pi){\,=\,}\mathbb E_\pi[\sum_{t\ge0}\gamma^t r(s_t,a_t)]\). The RND energy $\mathcal{E}(x)$ is used as a unified uncertainty signal  \(b^{+}(x)\): in online RL, it is added as an exploration bonus, \(\tilde r_t=r_t+\lambda_{\mathrm{coeff}}b^{+}(s_t)\); in offline RL, it is used as an anti-exploration penalty in the critic target, $r_t+\gamma\,\mathbb E_{a'}[Q_{\theta'}(s_{t+1},a')-\lambda_{\mathrm{coeff}}b^{+}(s_{t+1},a')]$. Here, \(x=s_t\) for online exploration and \(x=(s_t,a_t)\) for offline conservatism. Detailed RL configurations are provided in Appendix~\ref{sec:hyperparameters}.

% \paragraph{Hyperparameter choices for \texttt{RFF-RND}.}
We fix the output dimension to \(m=64\) for online RL and \(m=128\) for offline RL experiments, as performance stabilizes once \(m\) is sufficiently large. A detailed analysis on the effect of \(m\) is provided in Appendix~\ref{subsec:effect_of_m}. As with other RND-based methods, we calibrate the scale of the intrinsic reward by tuning \(\alpha\) and \(\lambda_{\mathrm{coeff}}\) on a validation grid and selecting dataset-specific values (see Appendix~\ref{subsec:alpha_lambda_performance}).

\begin{figure*}[t]
\centering
\setlength{\fboxsep}{0pt}

% =========================
% Row 1: Atari
% =========================
\begin{minipage}{0.93\textwidth}
\centering
\begin{minipage}{0.24\textwidth}
\centering
\includegraphics[width=\linewidth]{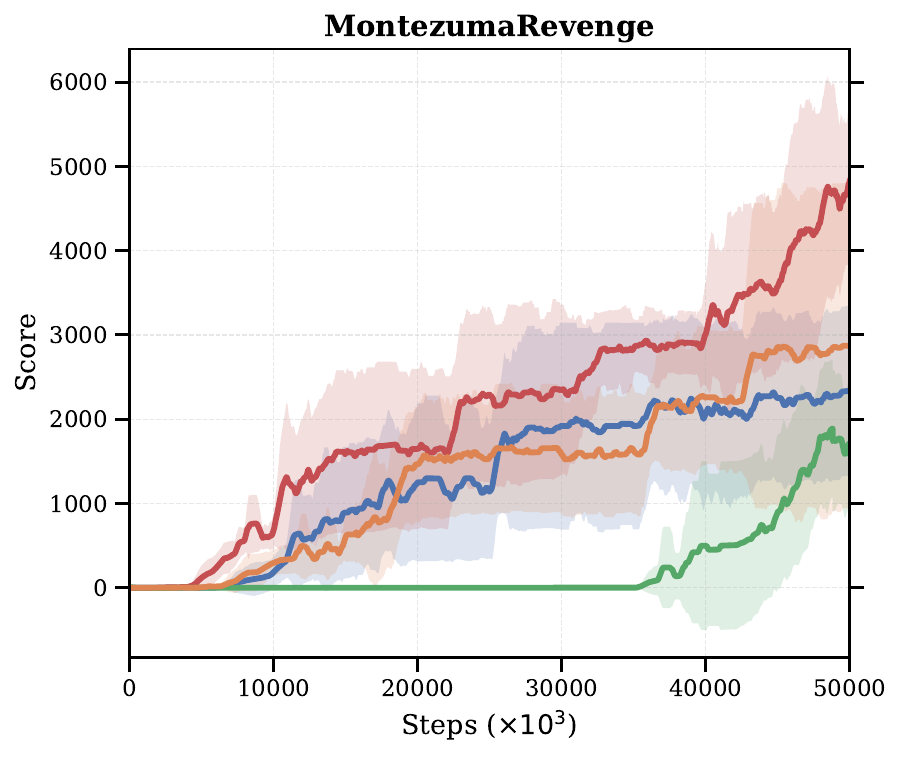}
\end{minipage}
\hfill
\begin{minipage}{0.24\textwidth}
\centering
\includegraphics[width=\linewidth]{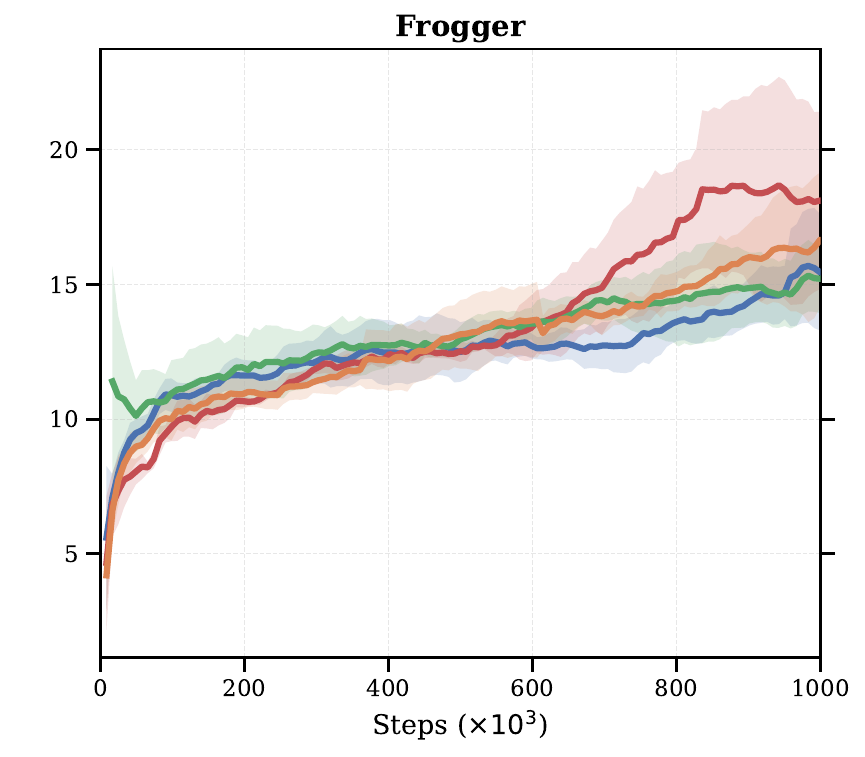}
\end{minipage}
\hfill
\begin{minipage}{0.24\textwidth}
\centering
\includegraphics[width=\linewidth]{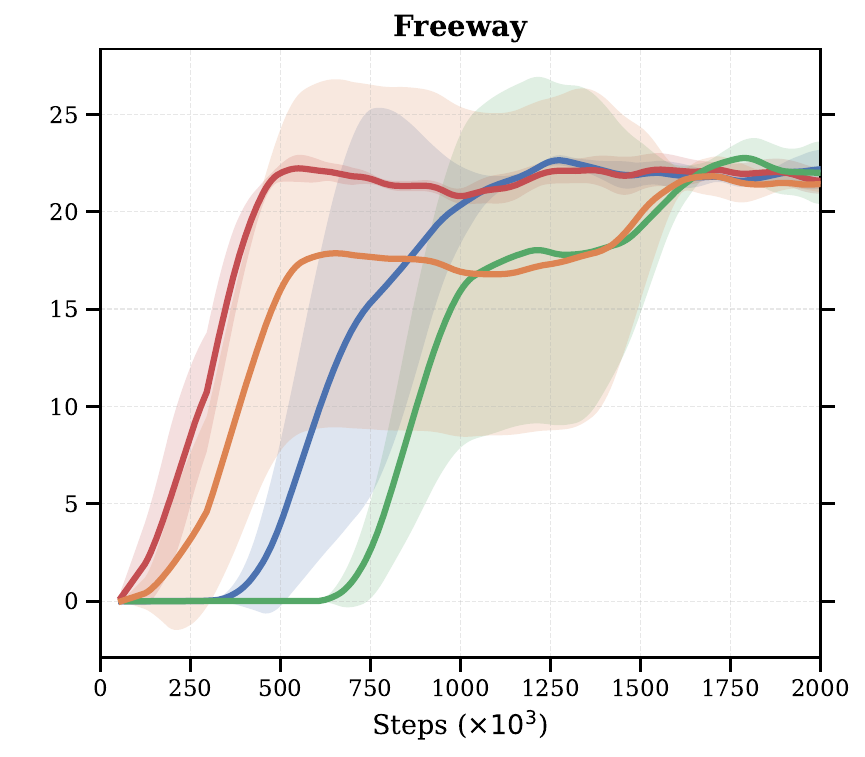}
\end{minipage}
\hfill
\begin{minipage}{0.25\textwidth}
\centering
\includegraphics[width=\linewidth]{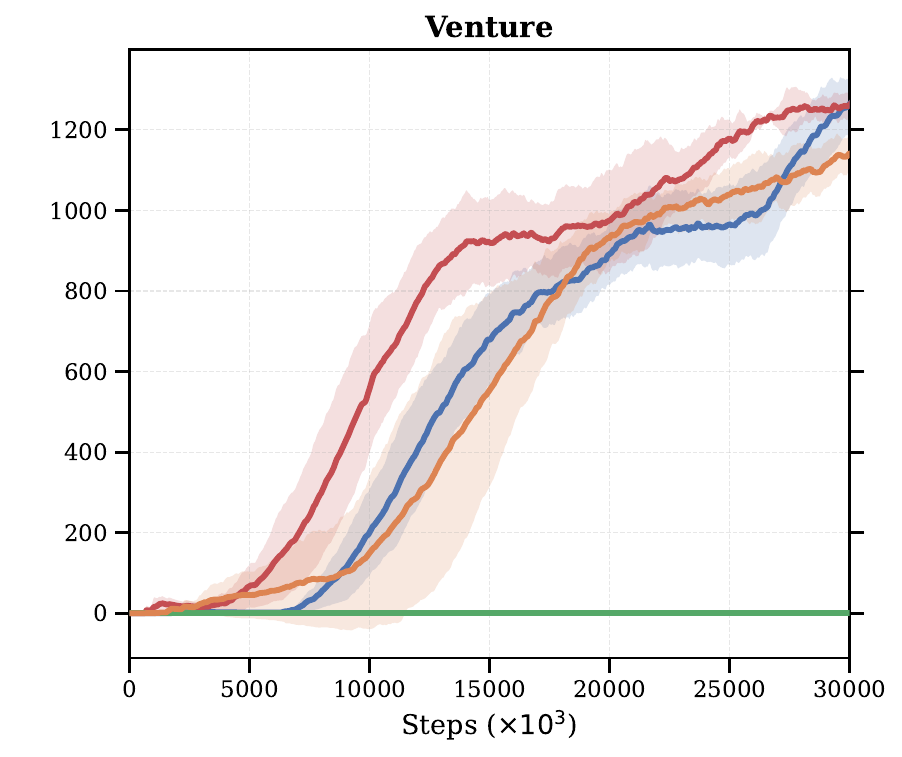}
\end{minipage}
\end{minipage}

\vspace{-0.8em}

% =========================
% Row 2: Fetch
% =========================
\begin{minipage}{0.93\textwidth}
\centering
\begin{minipage}{0.24\textwidth}
\centering
\includegraphics[width=\linewidth]{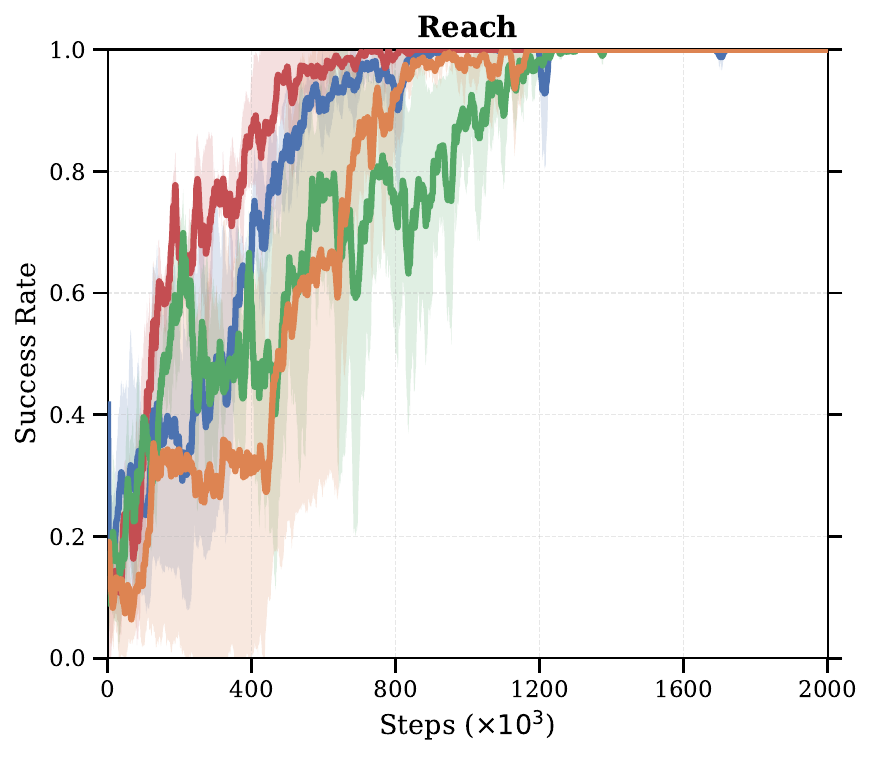}
\end{minipage}
\hfill
\begin{minipage}{0.24\textwidth}
\centering
\includegraphics[width=\linewidth]{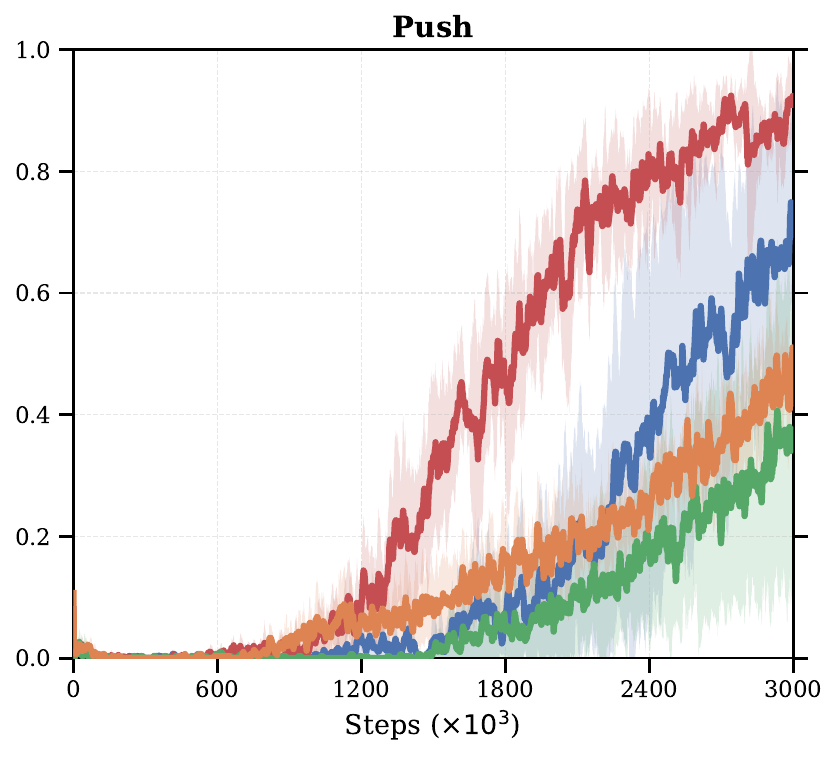}
\end{minipage}
\hfill
\begin{minipage}{0.24\textwidth}
\centering
\includegraphics[width=\linewidth]{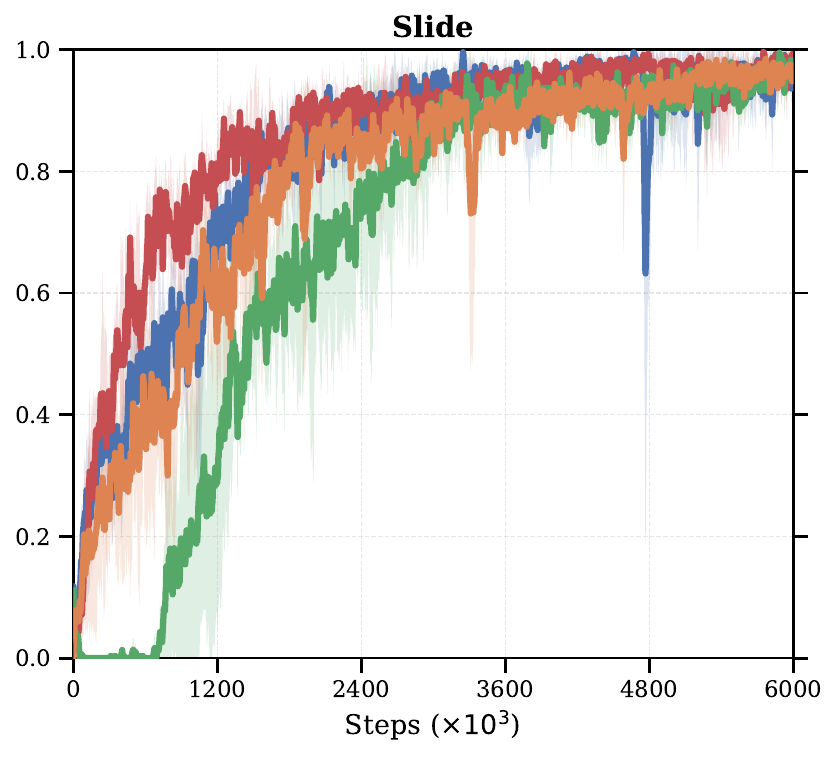}
\end{minipage}
\hfill
\begin{minipage}{0.25\textwidth}
\centering
\includegraphics[width=\linewidth]{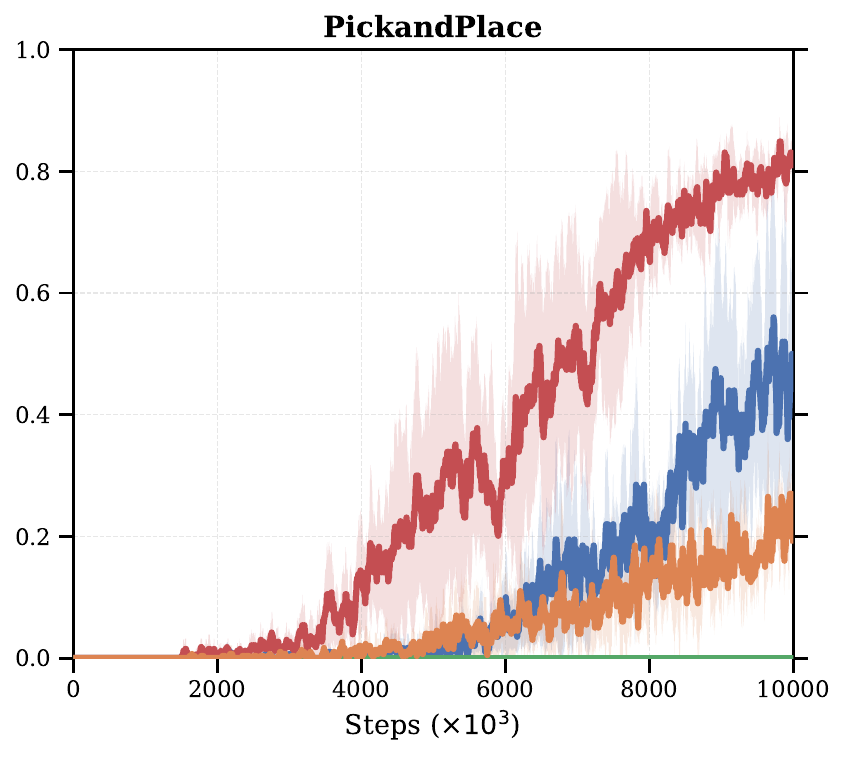}
\end{minipage}
\end{minipage}

\vspace{-0.2em}

% =========================
% Legend
% =========================
\begin{minipage}{1\textwidth}
\centering
\includegraphics[width=0.55\textwidth]{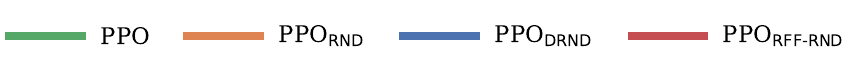}
\end{minipage}

\vspace{-0.8em}

\caption{\textbf{Mean episode return on online RL tasks.}
Top: Atari tasks (\emph{MontezumaRevenge}, \emph{Frogger}, \emph{Freeway}, \emph{Venture}); Bottom: Fetch Robotics tasks (\emph{Reach}, \emph{Push}, \emph{Slide},
\emph{PickAndPlace}). Results are averaged over 5 random seeds.}
\label{fig:onlineRL_graph}
\end{figure*}

\paragraph{Online RL.}\label{subsec:online_RL}
% \textcolor{blue}{Environment description of online RL}

% For the online setting, w
We evaluate \texttt{RFF-RND} on four Atari exploration
games (\emph{MontezumaRevenge}, \emph{Frogger}, \emph{Freeway}, \emph{Venture})
and four Fetch manipulation tasks (\emph{Reach}, \emph{Push}, \emph{Slide},
\emph{PickAndPlace}). We use PPO~\citep{schulman2017ppo} as the base algorithm
and compare PPO, PPO$_{\text{RND}}$~\citep{burda2018rnd},
PPO$_{\text{DRND}}$~\citep{yang2024exploration}, and the proposed
PPO$_{\text{\texttt{RFF-RND}}}$, isolating the effect of the intrinsic reward.

As shown in Figure~\ref{fig:onlineRL_graph}, PPO$_{\text{\texttt{RFF-RND}}}$
generally improves sample efficiency and achieves the best or competitive final
performance. On sparse-reward Atari tasks, the gains are most visible on
\emph{MontezumaRevenge}, where PPO$_{\text{\texttt{RFF-RND}}}$ continues to
improve while other methods plateau, and on \emph{Venture} and \emph{Freeway},
where it improves earlier during training. On \emph{Frogger}, 
% the advantage is more modest, but 
PPO$_{\text{\texttt{RFF-RND}}}$ remains consistently
competitive with other RND-based variants.
The Fetch tasks show a similar trend in goal-conditioned sparse-reward
manipulation. PPO$_{\text{\texttt{RFF-RND}}}$ solves \emph{Reach} fastest,
is the only method to reach \(80\%\) success on \emph{Push} within the training
budget, and achieves the best final performance on \emph{PickAndPlace}. On
\emph{Slide}, all methods approach comparable final performance, but
PPO$_{\text{\texttt{RFF-RND}}}$ learns faster early in training. Overall, these
results indicate that the proposed target-side design yields an intrinsic reward
that is useful for exploration across various online RL tasks.

\begin{table*}[t]
\centering
% \caption{Normalized scores for offline D4RL Locomotion tasks are reported as mean $\pm$ std, averaged over 10 random seeds. Each evaluation uses 100 episodes. Baseline results are taken from prior work. SAC and CQL from \citet{an2021uncertaintybased}, IQL, ReBRAC, and SAC$_{\text{RND}}$ from \citet{tarasov2023revisiting}, and SAC$_{\text{DRND}}$ from \citet{yang2024exploration}. \textbf{Bold} indicates best, \underline{underline} second-best.}
\caption{Normalized D4RL Locomotion scores are reported as mean \(\pm\) std over 10 seeds, with 100 evaluation episodes per seed. Baselines are taken from prior
work~\citep{an2021uncertaintybased,tarasov2023revisiting,yang2024exploration}.
\textbf{Bold} and \underline{underline} denote the best and second-best results.}
\vspace{-0.5em}
\label{tab:offline_rl_benchmark}

\setlength{\tabcolsep}{3pt}
\renewcommand{\arraystretch}{1.1}
\resizebox{\linewidth}{!}{%
\begin{tabular}{lccccccc}
\hline
Dataset & SAC & CQL & IQL & ReBRAC & SAC$_{\text{RND}}$ & SAC$_{\text{DRND}}$ & SAC$_{\text{\texttt{RFF-RND}}}$ \\
\hline
halfcheetah-random
 & $29.7 \pm 1.4$ & $\boldsymbol{31.1 \pm 3.5}$ & $19.5 \pm 0.8$ & $29.5 \pm 1.5$ & $27.6 \pm 2.1$ & $\underline{30.4 \pm 4.0}$ & $29.3 \pm 1.7$ \\
halfcheetah-medium
 & $55.2 \pm 27.8$ & $46.9 \pm 0.4$ & $50.0 \pm 0.2$ & $65.6 \pm 1.0$ & $66.4 \pm 1.4$ & $\boldsymbol{68.3 \pm 0.2}$ & $\underline{67.0 \pm 0.5}$ \\
halfcheetah-expert
 & $-0.8 \pm 1.8$ & $97.3 \pm 1.1$ & $95.5 \pm 2.1$ & $105.9 \pm 1.7$ & $102.6 \pm 4.2$ & $\underline{106.2 \pm 3.7}$ & $\boldsymbol{109.2 \pm 2.0}$ \\
halfcheetah-medium-expert
 & $28.4 \pm 19.4$ & $95.0 \pm 1.4$ & $92.7 \pm 2.8$ & $101.1 \pm 5.2$ & $107.6 \pm 2.8$ & $\underline{108.5 \pm 1.1}$ & $\boldsymbol{109.7 \pm 0.3}$ \\
halfcheetah-medium-replay
 & $0.8 \pm 1.0$ & $45.3 \pm 0.3$ & $42.1 \pm 3.6$ & $51.0 \pm 0.8$ & $51.2 \pm 3.2$ & $\underline{52.1 \pm 4.8}$ & $\boldsymbol{57.0 \pm 2.0}$ \\
halfcheetah-full-replay
 & $\boldsymbol{86.8 \pm 1.0}$ & $76.9 \pm 0.9$ & $75.0 \pm 0.7$ & $82.1 \pm 1.1$ & $81.2 \pm 1.3$ & $81.4 \pm 1.7$ & $\underline{84.3 \pm 1.1}$ \\
\hline
hopper-random
 & $9.9 \pm 1.5$ & $5.3 \pm 0.6$ & $10.1 \pm 5.9$ & $8.1 \pm 2.4$ & $19.6 \pm 12.4$ & $\boldsymbol{32.7 \pm 0.4}$ & $\underline{31.4 \pm 0.1}$ \\
hopper-medium
 & $0.8 \pm 0.0$ & $61.9 \pm 6.4$ & $65.2 \pm 4.2$ & $\underline{102.0 \pm 1.0}$ & $91.1 \pm 10.1$ & $98.5 \pm 1.1$ & $\boldsymbol{102.3 \pm 1.5}$ \\
hopper-expert
 & $0.7 \pm 0.0$ & $106.5 \pm 9.1$ & $108.8 \pm 3.1$ & $100.1 \pm 8.3$ & $109.7 \pm 0.5$ & $\underline{109.7 \pm 0.3}$ & $\boldsymbol{110.2 \pm  0.8}$ \\
hopper-medium-expert
 & $0.7 \pm 0.0$ & $96.9 \pm 15.1$ & $85.5 \pm 29.7$ & $107.0 \pm 6.4$ & $\underline{109.8 \pm 0.6}$ & $108.7 \pm 0.5$ & $\boldsymbol{112.0 \pm 0.2}$ \\
hopper-medium-replay
 & $7.4 \pm 0.5$ & $86.3 \pm 7.3$ & $89.6 \pm 13.2$ & $98.1 \pm 5.3$ & $97.2\pm 9.0$ & $\underline{100.5 \pm 1.0}$ & $\boldsymbol{102.7 \pm 0.4}$ \\
hopper-full-replay
 & $41.1 \pm 17.9$ & $101.9 \pm 0.6$ & $104.4 \pm 10.8$ & $107.1 \pm 0.4$ & $107.4 \pm 0.8$ & $\underline{108.2 \pm 0.7}$ & $\boldsymbol{109.6 \pm 0.4}$ \\

\hline
walker2d-random
 & $0.9 \pm 0.8$ & $5.4 \pm 1.7$ & $11.3 \pm 7.0$ & $21.5 \pm 0.1$ & $18.7 \pm 6.9$ & $\underline{21.7 \pm 0.1}$ & $\boldsymbol{22.0 \pm 1.4}$  \\
walker2d-medium
 & $-0.3 \pm 0.2$ & $79.5 \pm 3.2$ & $80.7 \pm 3.4$ & $82.5 \pm 3.6$ & $91.6 \pm 2.8$ & $\boldsymbol{95.2 \pm 0.7}$ & $\underline{94.3 \pm 0.9}$ \\
walker2d-expert
 & $0.7 \pm 0.3$ & $109.3 \pm 0.1$ & $96.9 \pm 32.3$ & $112.3 \pm 0.2$ & $104.5 \pm 22.8$ & $\boldsymbol{114.0 \pm 0.5}$ & $\underline{112.8 \pm 0.5}$ \\
walker2d-medium-expert
 & $1.9 \pm 3.9$ & $109.1 \pm 0.2$ & $\underline{112.1 \pm 0.5}$ & $111.6 \pm 0.3$ & $104.6 \pm 11.2$ & $109.6 \pm 1.0$ & $\boldsymbol{113.2 \pm 0.6}$ \\
walker2d-medium-replay
 & $-0.4 \pm 0.3$ & $76.8 \pm 10.0$ & $75.4 \pm 9.3$ & $77.3 \pm 7.9$ & $88.7 \pm 7.7$ & $\underline{91.0 \pm 2.9}$ & $\boldsymbol{101.7 \pm 3.7}$ \\
walker2d-full-replay
 & $27.9 \pm 47.3$ & $94.2 \pm 1.9$ & $97.5 \pm 1.4$ & $102.2 \pm 1.7$ & $105.3 \pm 3.2$ & $\underline{109.6 \pm 0.7}$ & $\boldsymbol{111.7 \pm 2.8}$ \\
\hline
average score
 & $16.2$ & $73.6$ & $72.9$ & $81.4$ & $82.5$ & $\underline{85.9}$ & $\textbf{87.8}$ \\
\hline
\end{tabular}
}
\vspace{-0.5em}
\end{table*}
\setlength{\tabcolsep}{2.5pt}
\renewcommand{\arraystretch}{1.08}

\begin{table*}[t]
\centering
% \caption{Normalized D4RL AntMaze scores are reported as mean \(\pm\) std, averaged over 10 random seeds. Each evaluation uses 100 episodes. Baselines are from prior work: SAC, CQL, IQL, and SAC$_{\text{RND}}$ from \citet{nikulin2023anti}, ReBRAC from \citet{tarasov2023revisiting}, and
% SAC$_{\text{DRND}}$ from \citet{yang2024exploration}. \textbf{Bold} indicates best, \underline{underline} second-best.}

\caption{
Normalized D4RL AntMaze scores are reported as mean \(\pm\) std over 10 seeds,
with 100 evaluation episodes per seed. Baselines are from prior
work~\citep{nikulin2023anti,tarasov2023revisiting,yang2024exploration}.
\textbf{Bold} and \underline{underline} denote the best and second-best results.
}
\vspace{-0.5em}
\label{tab:antmaze_offline_rl_benchmark}
\footnotesize
\resizebox{\linewidth}{!}{%
\begin{tabular}{lccccccc}
\hline
Dataset & SAC & CQL & IQL & ReBRAC & SAC$_{\text{RND}}$ & SAC$_{\text{DRND}}$ & SAC$_{\text{\texttt{RFF-RND}}}$ \\
\hline
antmaze-umaze 
 & 0.0 & 74.0 & 83.3 $\pm$ 4.5 & \textbf{97.8 $\pm$ 1.0} & 97.0 $\pm$ 1.5 & 95.8 $\pm$ 2.4 & \underline{97.0 $\pm$ 0.9} \\
antmaze-umaze-diverse 
 & 0.0 & 84.0 & 70.6 $\pm$ 3.7 & \underline{88.3 $\pm$ 13.0} & 66.0 $\pm$ 25.0 & 87.2 $\pm$ 3.2 & \textbf{88.4 $\pm$ 3.9} \\
antmaze-medium-play 
 & 0.0 & 61.2 & 64.6 $\pm$ 4.9 & 84.0 $\pm$ 4.2 & 38.5 $\pm$ 29.4 & \textbf{86.2 $\pm$ 5.4} & \underline{85.2 $\pm$ 6.1} \\
antmaze-medium-diverse 
 & 0.0 & 53.7 & 61.7 $\pm$ 6.1 & 76.3 $\pm$ 13.5 & 74.7 $\pm$ 10.7 & \textbf{83.0 $\pm$ 3.8} &  \underline{80.5 $\pm$ 5.5}\\
antmaze-large-play 
 & 0.0 & 15.8 & 42.5 $\pm$ 6.5 & \textbf{60.4 $\pm$ 26.1} & 43.9 $\pm$ 29.2 & 53.2 $\pm$ 4.1 & \underline{57.1 $\pm$ 8.2} \\
antmaze-large-diverse 
 & 0.0 & 14.9 & 27.6 $\pm$ 7.8 & \textbf{54.4 $\pm$ 25.1} & 45.7 $\pm$ 28.5 & 50.8 $\pm$ 10.5 & \underline{51.6 $\pm$ 9.8} \\
\hline
average score 
 & 0.0 & 50.6 & 58.4 & \textbf{76.9} & 61.0 & 76.0 & \underline{76.6} \\
\hline
\end{tabular}
}
\vspace{-1.5em}
\end{table*}

\paragraph{Offline RL.}\label{subsec:offline_RL}
% For the offline setting, w
We evaluate our method on D4RL locomotion tasks (\emph{HalfCheetah, Hopper, Walker2d})~\citep{fud4rl}, and additionally report results on \emph{AntMaze} tasks to assess performance in goal-directed exploration settings. We compare against strong ensemble-free offline RL baselines, including SAC~\citep{haarnoja2018sac}, CQL~\citep{kumar2020cql}, IQL~\citep{kostrikov2021iql}, ReBRAC~\citep{tarasov2023revisiting}, and RND-based variants SAC$_{\text{RND}}$~\citep{burda2018rnd} and SAC$_{\text{DRND}}$~\citep{yang2024exploration}. All RND-based variants are evaluated under the same SAC backbone and D4RL evaluation protocol to ensure fair comparisons.

As shown in Table~\ref{tab:offline_rl_benchmark},
SAC$_{\text{\texttt{RFF-RND}}}$ achieves the best average score on D4RL
locomotion among ensemble-free baselines, with an average score of \(87.8\).
It obtains the best or second-best performance on 17 out of 18 datasets, with
particularly strong gains on mixed-quality datasets such as
\emph{medium-replay}, \emph{full-replay}, and \emph{medium-expert}. These
results suggest that \texttt{RFF-RND} provides a support-aware uncertainty
penalty that helps exploit high-return behavior while discouraging poorly
supported actions,
% . This enables 
letting the actor more reliably exploit high-quality behavior already in the offline dataset.

Table~\ref{tab:antmaze_offline_rl_benchmark} reports additional results on
AntMaze. SAC$_{\text{\texttt{RFF-RND}}}$ remains competitive on these
goal-reaching tasks, achieving an average score of \(76.6\), close to the best
ensemble-free baseline. This suggests that the proposed uncertainty penalty can
extend beyond locomotion domains, although further adaptation to sparse-reward,
long-horizon navigation remains an interesting direction.

\section{Conclusion and Future Direction}
\label{sec:conclusion}

In this paper, we rigorously analyzed random network distillation (RND) overgeneralization under a GP target model and an NTK-regime predictor, showing that the RND score is a cross-kernel residual variance determined by the interaction between target covariance and predictor interpolation. This reveals residual collapse as a spectral under-excitation problem, where the target fails to excite modes in which predictor residuals can survive. Motivated by this mechanism, we proposed \texttt{RFF-RND}, a bandwidth-controlled target design that makes the target spectrum and spatial novelty scale explicit. Diagnostics and RL experiments show improved novelty discrimination and downstream performance, highlighting target geometry as a central design choice for reliable RND.

The theoretical framework proposed in this paper imposes no specific restrictions on the input domain. Specializing it to multiple input modalities, such as images, text, and time series, and analyzing the additional structures that arise in each case is therefore an interesting direction for future research.

\section{Practical Considerations for Spectral Target Design}
\label{sec:practical}

This section summarizes several practical issues discussed during the rebuttal
process that further clarify the scope of our spectral target design framework.
In particular, we consider data-normalized bandwidth selection, the interaction
between spectral design and stochastic nuisance variation, and the potential
role of anisotropic target spectra.

\paragraph{Data-normalized bandwidth selection.}
The RBF bandwidth $\alpha$ should be interpreted relative to the characteristic
scale of the input distribution. To obtain a simple data-dependent
normalization, we standardize each state and action dimension, concatenate them
into an input vector $x_i$, and compute the median pairwise distance
\[
r_{\mathrm{med}}
=
\operatorname{median}_{i\neq j}\|x_i-x_j\|_2 .
\]
We then parameterize the bandwidth as
\[
\alpha = c\,r_{\mathrm{med}}, \qquad c>0.
\]
For the RBF kernel
\[
k_\alpha(x,x')
=
\exp\left(
-\frac{\|x-x'\|_2^2}{2\alpha^2}
\right),
\]
two samples separated by $r_{\mathrm{med}}$ have similarity
$\exp(-1/(2c^2))$. Thus, $c$ directly controls the decay of target
correlation relative to the characteristic geometric scale of the dataset,
rather than its raw coordinate scale.

We evaluated this normalization on the 17 offline locomotion datasets used in
our experiments. The single choice $c=1/2$ yielded performance comparable to
the best grid-searched bandwidth on 10 datasets, while the candidate sets
$c\in\{1/2,1/3\}$ and $c\in\{1/2,1/3,1/5\}$ covered 14 and all 17 datasets,
respectively. These results suggest the compact data-normalized candidate set
\[
\alpha
\in
\left\{
r_{\mathrm{med}}/2,\;
r_{\mathrm{med}}/3,\;
r_{\mathrm{med}}/5
\right\},
\]
which substantially reduces the need for extensive task-specific bandwidth
search.

\paragraph{Stochastic nuisance variation and the noisy-TV problem.}
Spectral target design does not by itself eliminate the classical noisy-TV
failure mode. Our analysis predicts persistent novelty whenever target
variation overlaps with directions that remain difficult for the predictor,
regardless of whether such variation is task-relevant or arises from
irreducible stochasticity. Hence, allocating more target power to
predictor-hard directions should not be interpreted as a monotone preference
for arbitrarily high-frequency target spectra.

To make this trade-off explicit, consider an observation $x=(z,\xi)$, where
$z$ denotes the task-relevant state and
$\xi\sim\mathcal{N}(0,\sigma^2 I_q)$ is an independent nuisance variable.
For an isotropic RBF target kernel, two independent observations of the same
underlying state, $z=z'$, satisfy
\[
\mathbb{E}_{\xi,\xi'}
\left[
k_\alpha\big((z,\xi),(z,\xi')\big)
\right]
=
\left(
1+\frac{2\sigma^2}{\alpha^2}
\right)^{-q/2}.
\]
Therefore, when $\alpha$ is small relative to the nuisance scale, repeated
observations of the same underlying state can become nearly uncorrelated in
the target representation. If these nuisance directions are also difficult
for the predictor, their residual contribution persists and can be mistaken
for novelty. Conversely, increasing $\alpha$ suppresses stochastic variation
but may also smooth over task-relevant differences between states.

This conflict arises because a single isotropic bandwidth controls both types
of variation. When task-relevant and nuisance directions can be distinguished,
the same framework naturally admits an anisotropic target kernel,
\[
k_{\alpha_z,\alpha_\xi}
=
\exp\left(
-\frac{\|z-z'\|_2^2}{2\alpha_z^2}
-\frac{\|\xi-\xi'\|_2^2}{2\alpha_\xi^2}
\right).
\]
Here, $\alpha_\xi$ can be chosen large enough to suppress nuisance variation,
while $\alpha_z$ preserves sensitivity to task-relevant novelty. More broadly,
our analysis therefore suggests allocating target power to
\emph{task-relevant predictor-hard directions}, rather than indiscriminately
increasing target frequency. Identifying such directions from side
information, repeated observations, controlled augmentations, or invariant
representations provides a natural extension of the present isotropic
formulation.

\section*{Acknowledgement}
% This work was supported by the IITP grant funded by the Korean government(MSIT) in 2026 (RS-2022-II220951, Development of Uncertainty-Aware Agents Learning by Asking Questions).
This work was partly supported by the IITP grant funded by Korean government(MSIT) in 2026 (RS-2022-II220951, LBA, 50\%), and in part by the National Research Foundation of Korea (NRF) grant funded by the Korea government (MSIT) (RS-2026-25504174, 50\%).

\bibliographystyle{plainnat}
\bibliography{references}

\medskip

\small

\newpage
\input{checklist.tex}

\newpage
\appendix
\section{Contributions}
\label{app:contributions}
\begin{itemize}
\item \textbf{Minseok Jeong:} Project lead. Developed the core algorithms, conducted the mathematical analysis, and led manuscript preparation.
\item \textbf{Yechan Lee:} Led the downstream offline and online RL experiments.
\item \textbf{Hyewon Choi:} Supported the downstream offline and online RL experiments.
\item \textbf{Jeongyong Yang:} Contributed to manuscript writing and refinement of the technical exposition.
\item \textbf{SooJean Han:} Advised on the project and contributed to manuscript writing.
\end{itemize}

\section{Proofs}
\label{app:theory_proof}

\subsection{Proof of Theorem~\ref{thm:main}}
\label{app:proof_main}

% \begin{proof}
Fix an arbitrary query point \(x\in\mathcal X\). By
Assumption~\ref{asm:ntk_predictor}, the limiting NTK-regime predictor is
\[
f_p^\infty(x)
=
F_X^\top A_x^\top,
\qquad
A_x
=
\Theta_p(x,X)(\Theta_p+\lambda I)^{-1}.
\]
Let \(f_t^{(j)}\) denote the \(j\)-th coordinate of the target function, and let
\[
F_X^{(j)}
=
[f_t^{(j)}(x_1),\ldots,f_t^{(j)}(x_n)]^\top
\]
be the \(j\)-th column of \(F_X\). Under the GP target model, the output
coordinates are independent and identically distributed as
\[
f_t^{(j)}\sim \mathcal{GP}(0,k_t),
\qquad j=1,\ldots,m.
\]
Therefore, for each coordinate \(j\),
\[
\begin{bmatrix}
f_t^{(j)}(x)\\
F_X^{(j)}
\end{bmatrix}
\sim
\mathcal N\left(
0,
\begin{bmatrix}
k_t(x,x) & k_t(x,X)\\
k_t(X,x) & K_t
\end{bmatrix}
\right),
\]
where \(K_t=k_t(X,X)\).

The \(j\)-th coordinate of the residual is
\[
r^{(j)}(x)
\triangleq
f_t^{(j)}(x)-f_p^{\infty,(j)}(x)
=
f_t^{(j)}(x)-A_xF_X^{(j)}.
\]
Thus, conditional on \(X\) and the predictor kernel \(\Theta_p\), the same
deterministic interpolation weights \(A_x\) are applied to each target
coordinate. Since \(r^{(j)}(x)\) is a linear transformation of a jointly
Gaussian vector, it is Gaussian. Since the vector is centered, \(r^{(j)}(x)\) is
also centered, and its second moment is its variance:
\[
\mathbb E_{f_t}\!\left[\bigl(r^{(j)}(x)\bigr)^2\right]
=
\operatorname{Var}\!\left(f_t^{(j)}(x)-A_xF_X^{(j)}\right).
\]
Expanding this variance gives
\[
\operatorname{Var}\!\left(f_t^{(j)}(x)-A_xF_X^{(j)}\right)
=
k_t(x,x)
-2A_xk_t(X,x)
+
A_xK_tA_x^\top  =
\sigma^2_{\mathrm{cross}}(x;\,k_t,\Theta_p),
\]
where the last equality follows from Definition~\ref{def:cross_kernel}. Hence,
under the GP target model,
\[
r^{(j)}(x)
\sim
\mathcal N\!\left(
0,\sigma^2_{\mathrm{cross}}(x;\,k_t,\Theta_p)
\right).
\]

For finite output dimension \(m\), the ideal NTK-regime RND energy is
\[
\mathcal E_m(x)
=
\frac{1}{m}\|f_t(x)-f_p^\infty(x)\|_2^2
=
\frac{1}{m}\sum_{j=1}^m \bigl(r^{(j)}(x)\bigr)^2.
\]
Because the target coordinates \(\{f_t^{(j)}\}_{j=1}^m\) are independent GP
samples and the same deterministic weights \(A_x\) are applied coordinate-wise,
the residual coordinates \(\{r^{(j)}(x)\}_{j=1}^m\) are i.i.d. centered Gaussian
random variables with variance
\(\sigma^2_{\mathrm{cross}}(x;\,k_t,\Theta_p)\). Therefore,
\(\{(r^{(j)}(x))^2\}_{j=1}^m\) are i.i.d. with finite mean
\(\sigma^2_{\mathrm{cross}}(x;\,k_t,\Theta_p)\). By the strong law of large
numbers,
\[
\mathcal E_m(x)
=
\frac{1}{m}\sum_{j=1}^m \bigl(r^{(j)}(x)\bigr)^2
\xrightarrow[m\to\infty]{\mathrm{a.s.}}
\sigma^2_{\mathrm{cross}}(x;\,k_t,\Theta_p).
\]
% \end{proof}

% Note that the Gaussianity assumption is quite strong for the proof, which uses only the second-order covariance structure.
% In fact, Gaussianity provides the variance interpretation and connects the model to the GP target interpretation.
% Later, in Section~\ref{sec:spectral-design}, we see the RFF targets match the desired covariance structure in expectation even though they are not GP sample paths.

\subsection{Proof of Lemma~\ref{cor:spectral}}
\label{app:proof_cor_spectral}

% \begin{proof}
By definition,
\[
\widehat F_X
=
[f_p^\infty(x_1),\ldots,f_p^\infty(x_n)]^\top
=
\Theta_p(\Theta_p+\lambda I)^{-1}F_X .
\]
Hence the training-set residual is
\[
R_X
=
F_X-\widehat F_X
=
\left(I-\Theta_p(\Theta_p+\lambda I)^{-1}\right)F_X .
\]
Using
\[
I-\Theta_p(\Theta_p+\lambda I)^{-1}
=
(\Theta_p+\lambda I)(\Theta_p+\lambda I)^{-1}
-
\Theta_p(\Theta_p+\lambda I)^{-1}
=
\lambda(\Theta_p+\lambda I)^{-1},
\]
we obtain
\[
R_X
=
\lambda(\Theta_p+\lambda I)^{-1}F_X .
\]

Now let \(\Theta_p=U_p\Lambda_pU_p^\top\). Then
\[
\lambda(\Theta_p+\lambda I)^{-1}
=
U_p
\operatorname{diag}\!\left(
\frac{\lambda}{\lambda_{p,1}+\lambda},
\ldots,
\frac{\lambda}{\lambda_{p,n}+\lambda}
\right)
U_p^\top .
\]
Therefore, for the \(i\)-th eigenvector \(u_i\),
\[
u_i^\top R_X
=
u_i^\top \lambda(\Theta_p+\lambda I)^{-1}F_X
=
\frac{\lambda}{\lambda_{p,i}+\lambda}u_i^\top F_X .
\]
Taking squared Euclidean norm gives
\[
\|u_i^\top R_X\|_2^2
=
\left(\frac{\lambda}{\lambda_{p,i}+\lambda}\right)^2
\|u_i^\top F_X\|_2^2 .
\]

Under the GP target model, the output coordinates are independent and each
column \(F_{X,j}\) satisfies
\[
F_{X}^{(j)}\sim \mathcal N(0,K_t).
\]
Thus, for each coordinate \(j\),
\[
\mathbb E\left[(u_i^\top F_{X}^{(j)})^2\right]
=
u_i^\top K_t u_i .
\]
Since \(F_X\in\mathbb R^{n\times m}\), summing over the \(m\) output coordinates yields
\[
\mathbb E\|u_i^\top F_X\|_2^2
=
\sum_{j=1}^m
\mathbb E\left[(u_i^\top F_{X}^{(j)})^2\right]
=
m\,u_i^\top K_tu_i .
\]
Combining the two displays,
\[
\mathbb{E}\|u_i^\top R_X\|_2^2
=
m
\left(\frac{\lambda}{\lambda_{p,i}+\lambda}\right)^2
u_i^\top K_t\,u_i .
\]
% This proves the claim.
% \end{proof}

% Since $R_X = [I - \Theta_p(\Theta_p+\lambda I)^{-1}]F_X
% = \lambda(\Theta_p+\lambda I)^{-1}F_X \triangleq B_\lambda F_X$,
% and $\mathbb{E}[F_X F_X^\top] = m \cdot K_t$,
% \[
% \mathbb{E}[R_X R_X^\top]
% = m \cdot B_\lambda K_t B_\lambda
% = m \cdot U_p D_\lambda U_p^\top K_t U_p D_\lambda U_p^\top,
% \]
% where $D_\lambda = \mathrm{diag}(\lambda/(\lambda_{p,i}+\lambda))$.
% Projecting onto $u_i$ gives the result.

\subsection{Proof of Corollary~\ref{cor:rff-energy}}
\label{app:proof_cor:rff-energy}

% \begin{proof}
Fix \(x\in\mathcal X\). Under the NTK-regime predictor, the prediction at \(x\)
is obtained by applying the same interpolation weights to each target coordinate:
\[
f_p^\infty(x)=F_X^\top A_x^\top,
\qquad
A_x=\Theta_p(x,X)(\Theta_p+\lambda I)^{-1}.
\]
Let \(F_X^{(j)}=[f_t^{(j)}(x_1),\ldots,f_t^{(j)}(x_n)]^\top\) denote the
\(j\)-th target coordinate on the dataset. The corresponding residual coordinate is
\[
r^{(j)}(x)
=
f_t^{(j)}(x)-A_xF_X^{(j)}.
\]
By the zero-mean and covariance assumptions,
\[
\mathbb E[f_t^{(j)}(x)^2]=k_t(x,x),\qquad
\mathbb E[F_X^{(j)}f_t^{(j)}(x)]=k_t(X,x),
\qquad
\mathbb E[F_X^{(j)}(F_X^{(j)})^\top]=K_t .
\]
Therefore,
\[
\begin{aligned}
\mathbb E\!\left[(r^{(j)}(x))^2\right]
&=
\mathbb E\!\left[
\left(f_t^{(j)}(x)-A_xF_X^{(j)}\right)^2
\right] \\
&=
k_t(x,x)
-2A_xk_t(X,x)
+A_xK_tA_x^\top \\
&=
\sigma^2_{\mathrm{cross}}(x;k_t,\Theta_p).
\end{aligned}
\]
Thus, for finite \(m\),
\[
\mathbb E[\mathcal E_m(x)]
=
\mathbb E\left[
\frac1m\sum_{j=1}^m (r^{(j)}(x))^2
\right]
=
\sigma^2_{\mathrm{cross}}(x;k_t,\Theta_p).
\]

If the target coordinates are i.i.d., then the random variables
\(\{(r^{(j)}(x))^2\}_{j=1}^m\) are also i.i.d. with finite mean
\(\sigma^2_{\mathrm{cross}}(x;k_t,\Theta_p)\). Hence, by the strong law of large
numbers,
\[
\mathcal E_m(x)
=
\frac1m\sum_{j=1}^m (r^{(j)}(x))^2
\xrightarrow[m\to\infty]{a.s.}
\sigma^2_{\mathrm{cross}}(x;k_t,\Theta_p).
\]

\subsection{Proof of Proposition~\ref{prop:frequency-residual}}
\label{app:proof_frequency_residual}

% \begin{proof}
Let \(\{u_\omega\}_\omega\) denote the discrete Fourier basis on the periodic grid. Since \(X\) is a uniform grid on a periodic domain and both kernels are shift-invariant, the corresponding Gram matrices are circulant. Hence they are simultaneously diagonalized by the discrete Fourier basis~\citep{gray2006toeplitz}.
\[
\Theta_p u_\omega
=
\widehat{\theta}_p(\omega) u_\omega,
\qquad
K_t^\alpha u_\omega
=
\widehat{k}_t^\alpha(\omega) u_\omega .
\]
Equivalently,
\[
\Theta_p
=
U_{\mathcal F}\widehat{\Theta}_p U_{\mathcal F}^*,
\qquad
K_t^\alpha
=
U_{\mathcal F}\widehat{K}_t^\alpha U_{\mathcal F}^*,
\]
where \(U_{\mathcal F}\) is the discrete Fourier matrix and \(\widehat{\Theta}_p\), \(\widehat{K}_t^\alpha\) are diagonal matrices whose diagonal entries are \(\widehat{\theta}_p(\omega)\) and
\(\widehat{k}_t^\alpha(\omega)\), respectively.

From Lemma~\ref{cor:spectral}, the training-set residual satisfies
\[
R_X
=
\lambda(\Theta_p+\lambda I)^{-1}F_X.
\]
Projecting this residual onto the Fourier mode \(u_\omega\) gives
\[
u_\omega^*R_X
=
u_\omega^*\lambda(\Theta_p+\lambda I)^{-1}F_X.
\]
Since \(u_\omega\) is an eigenvector of \(\Theta_p\),
\[
u_\omega^*\lambda(\Theta_p+\lambda I)^{-1}
=
\frac{\lambda}{\widehat{\theta}_p(\omega)+\lambda}u_\omega^*.
\]
Therefore,
\[
u_\omega^*R_X
=
\frac{\lambda}{\widehat{\theta}_p(\omega)+\lambda}
u_\omega^*F_X.
\]
Taking squared Euclidean norm over the \(m\) output coordinates,
\[
\|u_\omega^*R_X\|_2^2
=
\left(
\frac{\lambda}{\widehat{\theta}_p(\omega)+\lambda}
\right)^2
\|u_\omega^*F_X\|_2^2.
\]

Under the GP target model, each output coordinate satisfies
\[
F_X^{(j)}\sim\mathcal N(0,K_t^\alpha),
\qquad j=1,\ldots,m,
\]
independently. Hence, for each coordinate,
\[
\mathbb E\left[
\left|u_\omega^*F_X^{(j)}\right|^2
\right]
=
u_\omega^*K_t^\alpha u_\omega.
\]
Since \(u_\omega\) is also an eigenvector of \(K_t^\alpha\),
\[
u_\omega^*K_t^\alpha u_\omega
=
\widehat{k}_t^\alpha(\omega),
\]
where we use the normalization \(\|u_\omega\|_2=1\). Summing over the \(m\) output coordinates gives
\[
\mathbb E\|u_\omega^*F_X\|_2^2
=
m\,\widehat{k}_t^\alpha(\omega).
\]
Combining the above displays,
\[
\mathbb E\|u_\omega^*R_X\|_2^2
=
m
\left(
\frac{\lambda}{\widehat{\theta}_p(\omega)+\lambda}
\right)^2
\widehat{k}_t^\alpha(\omega).
\]
Therefore, the output-normalized residual contribution is
\[
r(\omega)
\triangleq
\frac{1}{m}\mathbb E\|u_\omega^*R_X\|_2^2
% \]
% is
% \[
% r(\omega)
=
\left(
\frac{\lambda}{\widehat{\theta}_p(\omega)+\lambda}
\right)^2
\widehat{k}_t^\alpha(\omega).
\]

\newpage
\section{Ablation on Target Kernel Structure}
\label{app:shift_invariant_kernels}
Our construction is not tied to a particular target kernel. By Bochner's
theorem, any shift-invariant kernel with spectral density \(p_\alpha\) can be
instantiated through random Fourier features, and is therefore compatible with
both our analysis and the \texttt{RFF-RND} construction. To test the sensitivity
to kernel choice, we compare four representative shift-invariant kernels:
RBF, Laplace, Mat\'ern-\(\tfrac{3}{2}\), and Mat\'ern-\(\tfrac{5}{2}\), all with the
same bandwidth parameter \(\alpha=0.5\). Writing \(r \triangleq
\|x-x'\|_2/\alpha\), these kernels are
\[
k_{\mathrm{RBF}}(x,x')
=
\exp\left(-\frac{r^2}{4}\right),
\]
\[
k_{\mathrm{Lap}}(x,x')
=
\exp(-r),
\]
\[
k_{\mathrm{Mat\acute ern}\text{-}3/2}(x,x')
=
\left(1+\sqrt{3}\,r\right)\exp(-\sqrt{3}\,r),
\]
\[
k_{\mathrm{Mat\acute ern}\text{-}5/2}(x,x')
=
\left(1+\sqrt{5}\,r+\frac{5}{3}r^2\right)\exp(-\sqrt{5}\,r).
\]

Empirically, the four kernel choices are difficult to distinguish in any
systematic or significant way. They produce broadly similar energy scales and
substantially sharper coverage discrimination than standard RND. This suggests
that the main benefit comes from enforcing a localized, distance-decaying target
covariance, rather than from a delicate choice of the specific shift-invariant
kernel family.

\begin{figure}[h]
    \centering
    \includegraphics[width=1.0\linewidth]{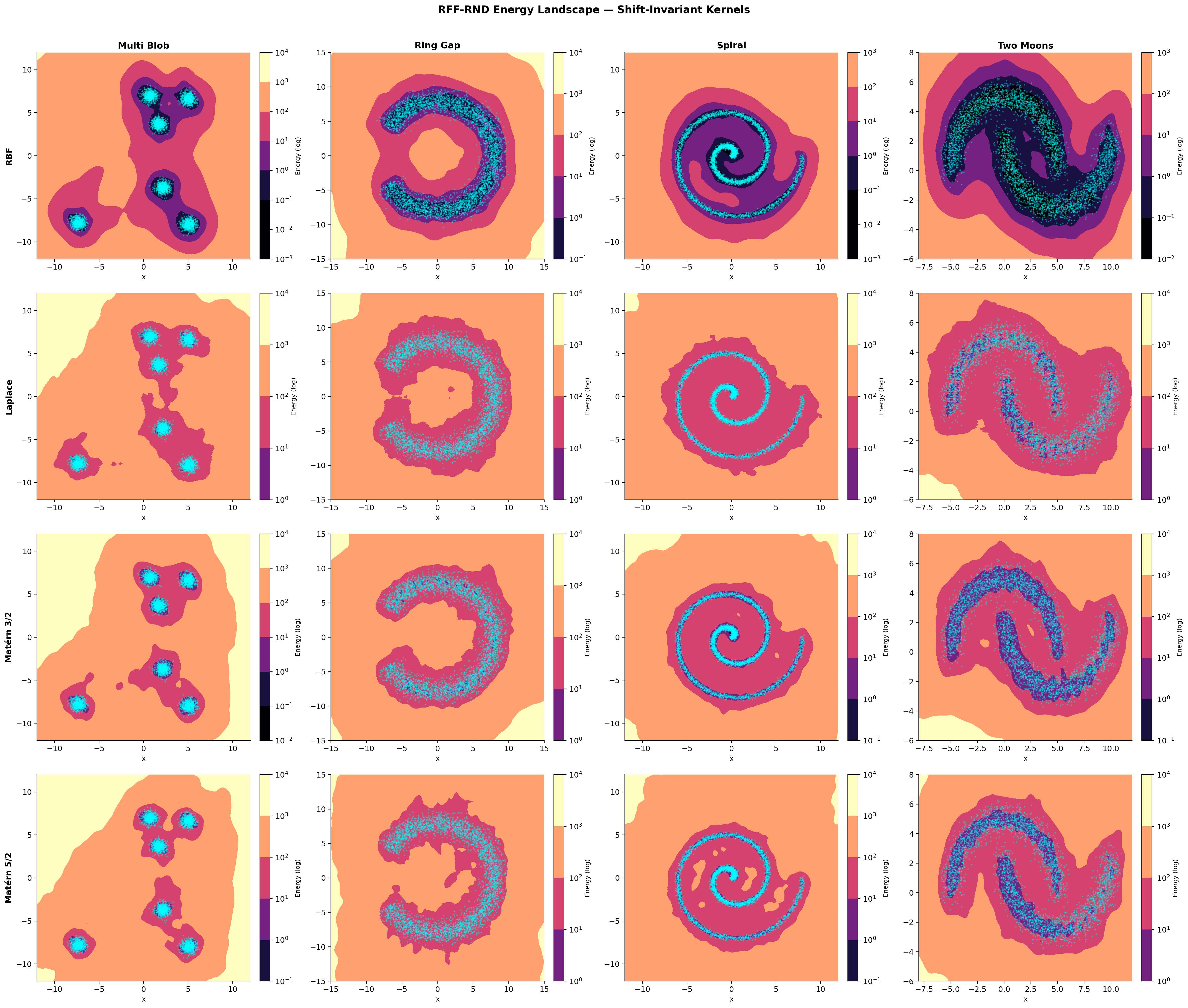}
    \caption{Comparison of \texttt{RFF-RND} using four shift-invariant target
    kernels---RBF, Laplace, Mat\'ern-\(\tfrac{3}{2}\), and Mat\'ern-\(\tfrac{5}{2}\)---all with the same bandwidth \(\alpha=0.5\). The four kernel choices show very
    similar behavior, with no significant qualitative difference in discrimination
    performance or overall energy scale. In contrast, all kernelized variants
    provide substantially clearer ID/OOD separation than standard RND.}
    \label{fig:rff_kernel_comparison}
\end{figure}

\newpage
% \section{Finite-Output Verification of Theorem~\ref{thm:main}}
\section{Diagnostics and Ablations for \texttt{RFF-RND}}
\subsection{Diagnostics on Target Output Dimension \texorpdfstring{$m$}{m} and Predictor Layer Width \texorpdfstring{$h$}{h}}
\label{app:ablation_rffrnd}
Corollary~\ref{cor:rff-energy} states that, under the RFF target and
NTK-regime predictor, the ideal infinite-output RND energy satisfies
\[
\mathcal{E}_{\infty}(x)
=
\sigma^2_{\rm cross}(x;\,k_t^\alpha,\Theta_p),
\quad \text{a.s.},
\]
for each fixed \(x\in\mathcal X\), where $\mathcal{E}_{\infty}(x)
=\lim_{m\to\infty}\mathcal{E}_m(x)$. In practice, however, RND uses a single realization of the target with finite output dimension \(m\). For a fixed query point \(x\), the residual
coordinates are i.i.d. under the RFF target construction, and hence
\[
\mathcal{E}_m(x)
=
\frac{1}{m}\|f_t(x)-f_p^\infty(x)\|_2^2
\xrightarrow[m\to\infty]{\mathrm{a.s.}}
\mathcal{E}_\infty(x)
=
\sigma^2_{\rm cross}(x;\,k_t^\alpha,\Theta_p),
\quad \text{pointwise in } x.
\]
Thus, the finite-output empirical energy \(\mathcal{E}_m(x)\) should
approach the theoretical prediction as \(m\) increases. We verify this
concentration empirically by measuring the agreement between
\(\mathcal{E}_m(x)\) and \(\sigma^2_{\rm cross}(x;\,k_t^\alpha,\Theta_p)\) as a
function of \(m\).

\paragraph{Protocol.}
We use the Two Moons dataset (\(n=150\), \(d=2\)). The predictor is a
2-hidden-layer MLP with hidden width \(h \in \{16,32,64\}\) and output
dimension \(m \in \{4,8,16,32,64, 128, 256\}\). The empirical NTK \(\Theta_p\) is
computed via full Jacobians at initialization. The theoretical value
\(\sigma^2_{\rm cross}(x;\,k_t^\alpha,\Theta_p)\) is computed in closed form using
Definition~\ref{def:cross_kernel}. The RFF target uses bandwidth
\(\alpha=0.5\) and \(\lambda=0.1\). Agreement is measured by Pearson \(r\)
between \(\sigma^2_{\rm cross}(x;\,k_t^\alpha,\Theta_p)\) and \(\mathcal{E}_m(x)\),
evaluated on a \(22\times 22\) grid covering the input space and averaged
over 5 random seeds.

\begin{figure}[h]
    \centering
    \includegraphics[width=0.95\linewidth]{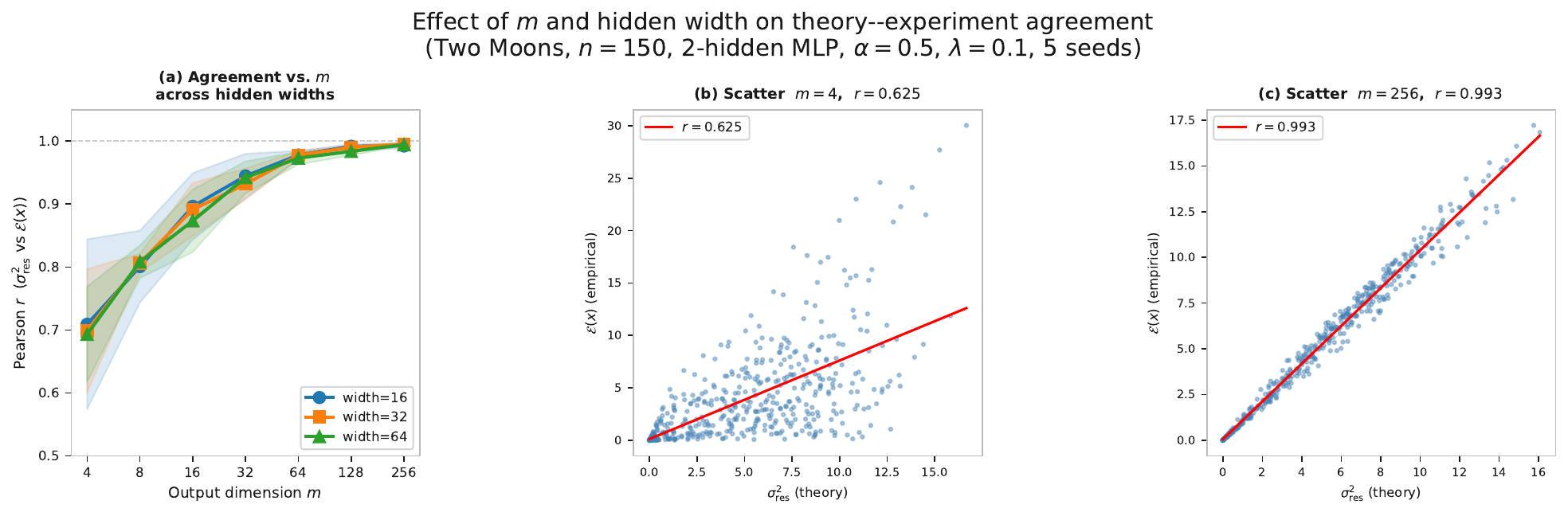}
    \caption{As the RFF target output dimension \(m\) increases, the finite-output
    empirical energy \(\mathcal{E}_m(x)\) concentrates around its infinite-output
    limit \(\mathcal{E}_\infty(x)=\sigma^2_{\rm cross}(x;\,k_t^\alpha,\Theta_p)\).
    Since the output coordinates are i.i.d., this concentration follows from
    the law of large numbers and yields higher empirical agreement with the
    theoretical prediction.}
    \label{fig:m_vs_r}
\end{figure}

% \paragraph{Agreement is robust to hidden width.}
\paragraph{Interpretation.}
Figure~\ref{fig:m_vs_r}-(a) shows Pearson \(r\) as a function of the RFF target output dimension \(m\) across three hidden widths. The three curves are nearly indistinguishable across \(h\in\{16,32,64\}\). This suggests that the observed agreement is robust to predictor width. On this low-dimensional two-moons dataset, even the narrow predictors induce sufficiently similar empirical NTK geometry for the residual-variance prediction to remain stable across widths. We note that in higher-dimensional settings, wider networks may be needed to obtain a reliable empirical NTK estimate; we leave this verification to future work.

% \paragraph{Agreement improves with \(m\).}
Figure~\ref{fig:m_vs_r}-(a) also shows the agreement improves consistently as \(m\) increases: starting from a relatively noisy finite-output
estimate at \(m=4\), the correlation rapidly approaches one and reaches nearly perfect agreement by \(m=256\). This trend is consistent with the concentration argument above, since \(\mathcal E_m(x)\) is an empirical average over \(m\)
independent target coordinates. The scatter plots make this effect explicit. At \(h=64, m=4\), substantial finite-output fluctuations remain and the empirical energy only coarsely tracks \(\sigma^2_{\rm cross}(x;\,k_t,\Theta_p)\) (Figure~\ref{fig:m_vs_r}(b)). In contrast, at \(h=64, m=256\), the scatter is almost linear with Pearson \(r=0.993\), showing that the finite-output realization closely matches the infinite-output prediction (Figure~\ref{fig:m_vs_r}-(c)).

% \paragraph{Interpretation.}
The key takeaway is that Corollary~\ref{cor:rff-energy} characterizes the infinite-output limit of NTK predictor RND energy, while practical \texttt{RFF-RND} uses a single finite-dimensional target realization. For small \(m\), \(\mathcal E_m(x)\) has noticeable finite-output variance and therefore only approximately follows its limit. Increasing \(m\) reduces this variance and makes a single \texttt{RFF-RND} realization track \(\sigma^2_{\rm cross}(x;\,k_t,\Theta_p)\) more reliably. In practice, the agreement already becomes strong at moderate dimensions, with \(m=64\) yielding high correlations across all hidden widths, and becomes nearly exact by \(m=256\).

\clearpage
\newpage

\subsection{Frequency-domain Diagnostic of Proposition \ref{prop:frequency-residual}}
\label{app:frequency_residual_verification}

Proposition~\ref{prop:frequency-residual} states that, on a uniform periodic
grid with shift-invariant predictor and target kernels, the residual power at
frequency \(\omega\) decomposes as
\[
r_\alpha(\omega)
=
\frac{1}{m}\mathbb E\|u_\omega^* R_X\|_2^2
=
\left(
\frac{\lambda}{\widehat{\Theta}_p(\omega)+\lambda}
\right)^2
\widehat{k}_t^\alpha(\omega).
\]
This experiment verifies the product-form decomposition in the idealized setting
where the proposition applies exactly. The goal is not to evaluate downstream
performance, but to visualize how the RFF target spectrum is filtered by the
predictor residual filter.

\paragraph{Protocol.}
We use a one-dimensional uniform periodic grid with \(N=512\) points on
\([0,1)\). The predictor kernel is a periodic RBF kernel with bandwidth
\(\alpha_p=0.08\), and the ridge parameter is \(\lambda=0.05\). We vary the target
RBF bandwidth over
\[
\alpha_t \in \{0.02,0.04,0.08,0.16,0.32\}.
\]
For each \(\alpha_t\), we compute the target spectrum
\(\widehat{k}_t^{\alpha_t}(\omega)\), the predictor residual filter
\[
\left(
\frac{\lambda}{\widehat{\Theta}_p(\omega)+\lambda}
\right)^2,
\]
and the theoretical residual spectrum
\[
r_{\alpha_t}^{\mathrm{theory}}(\omega)
=
\left(
\frac{\lambda}{\widehat{\Theta}_p(\omega)+\lambda}
\right)^2
\widehat{k}_t^{\alpha_t}(\omega).
\]
To obtain the empirical residual spectrum, we sample target functions from the
periodic GP with covariance \(k_t^{\alpha_t}\), apply the NTK residual operator,
and measure the residual power in each Fourier mode. The empirical curves are
averaged over Monte Carlo target samples.

\begin{figure}[h]
    \centering
    \includegraphics[width=1.0\linewidth]{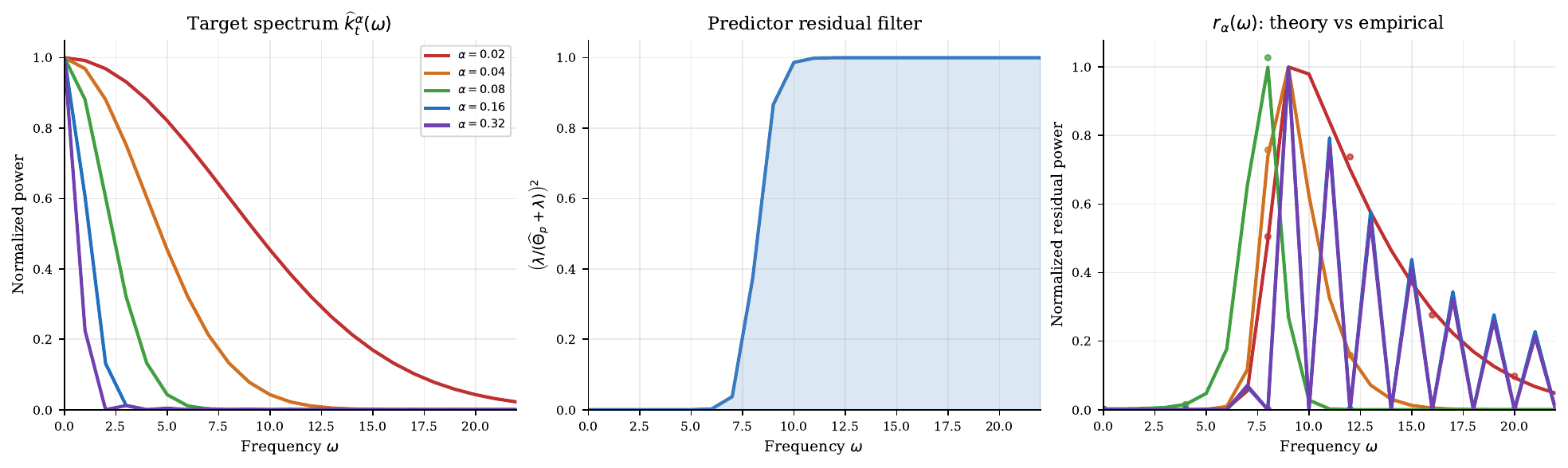}
    \caption{
    \textbf{Frequency-domain residual decomposition on a periodic grid.}
    Left: changing the target bandwidth \(\alpha_t\) reshapes the target spectrum
    \(\widehat{k}_t^{\alpha_t}(\omega)\). Middle: the predictor kernel induces a
    residual filter
    \((\lambda/(\widehat{\Theta}_p(\omega)+\lambda))^2\), which suppresses
    frequencies with large predictor spectral mass. Right: the residual spectrum
    is the product of these two terms. Empirical residuals obtained by sampling
    target functions and applying the NTK residual operator follow the predicted
    frequency-wise structure.
    }
    \label{fig:frequency_residual}
\end{figure}

\paragraph{Interpretation.}
Figure~\ref{fig:frequency_residual} illustrates that residual signal does not
appear wherever the target has large spectral power. Low frequencies have large
target power for smooth targets, but they are strongly suppressed by the
predictor residual filter. Conversely, very high frequencies pass the filter but
may receive little target power. Residual power is therefore largest where the
target spectrum overlaps with frequencies that are not interpolated away by the
predictor. This supports the interpretation of the RFF bandwidth as a spectral
knob: changing the target bandwidth moves target power across frequencies and
thereby controls where RND residual signal can survive.

\newpage

\subsection{Ablation on Energy Landscapes Across \texorpdfstring{\(\alpha\)}{a}, \texorpdfstring{\(m\)}{m}, and \texorpdfstring{\(h\)}{h}}
\label{app:ablation_energy}
% \section{Effects of Bandwidth $\alpha$, Output Dimension $m$, and Predictor Width $h$}

We further examine how the \texttt{RFF-RND} energy landscape changes with the
target bandwidth \(\alpha\), target output dimension \(m\), and predictor width
\(h\). These ablations are intended to visualize the geometric effect of each
design parameter, rather than to tune downstream RL performance.

\begin{figure*}[h]
    \centering
    \captionsetup{skip=-1pt}
    \includegraphics[width=1.0\linewidth]{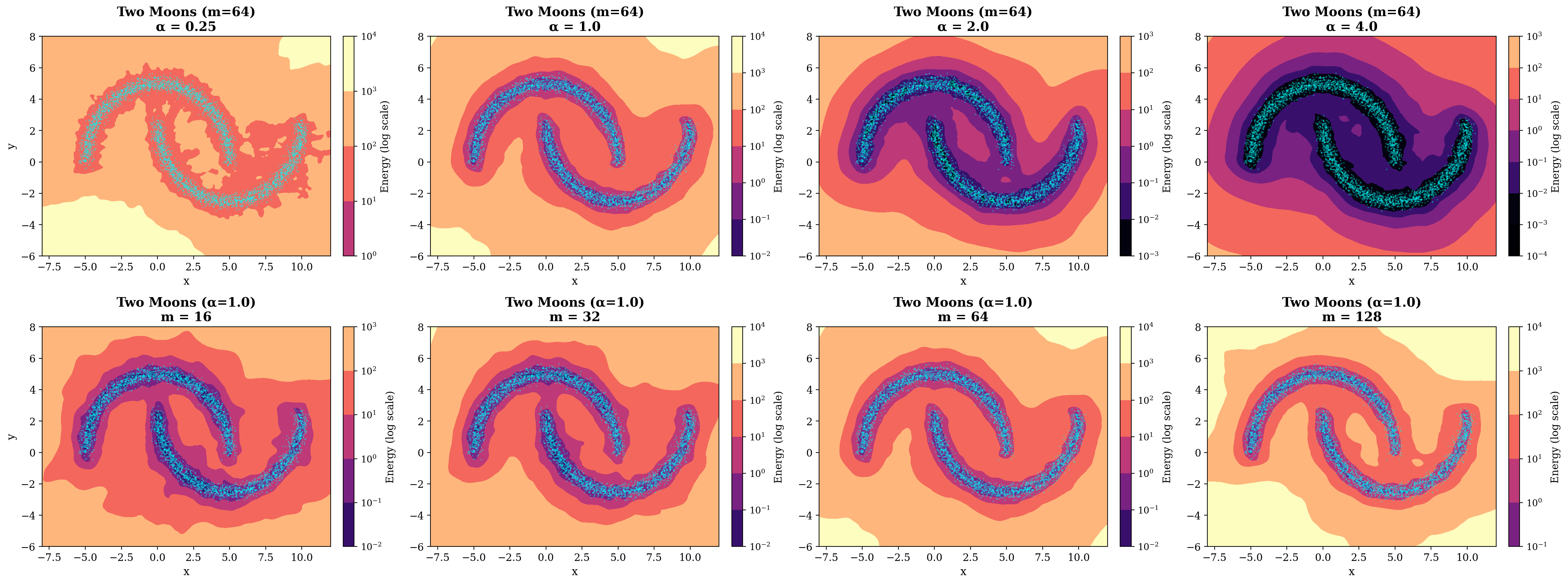}
    \caption{Energy landscapes of \texttt{RFF-RND} on the Two Moons dataset
    % (log scale). 
    over varying $\alpha$ and $m$.
    \textbf{(Top Row)} Varying the RBF length-scale $\alpha$ with $m=64$ fixed. \textbf{(Bottom Row)} Varying the RFF feature dimension $m$ with $\alpha=1.0$ fixed. 
    Overall, smaller $\alpha$ yields more localized (but noisier) energy bands around the manifold, whereas larger $\alpha$ over-smooths the landscape and blurs manifold boundaries. Increasing $m$ improves the kernel approximation and stabilizes the energy, producing smoother and more consistent manifold separation.}
    \label{fig:rnd_energy_heatmap_alpha_vs_dim}
\end{figure*}

\paragraph{Effect of bandwidth \(\alpha\) and output dimension \(m\).}
Figure~\ref{fig:rnd_energy_heatmap_alpha_vs_dim} shows energy landscapes on the Two Moons dataset. Varying \(\alpha\) changes the spatial scale of the target kernel: smaller \(\alpha\) produces more localized energy around the sampled manifold, while larger \(\alpha\) over-smooths the landscape and blurs the separation between the manifold and nearby off-manifold regions. Varying \(m\) primarily affects finite-output stability. Small \(m\) yields noisier energy landscapes,
whereas larger \(m\) gives smoother and more consistent separation, consistent with the concentration behavior in Corollary~\ref{cor:rff-energy}.

\begin{figure}[h]
    \centering
    \includegraphics[width=1.0\linewidth]{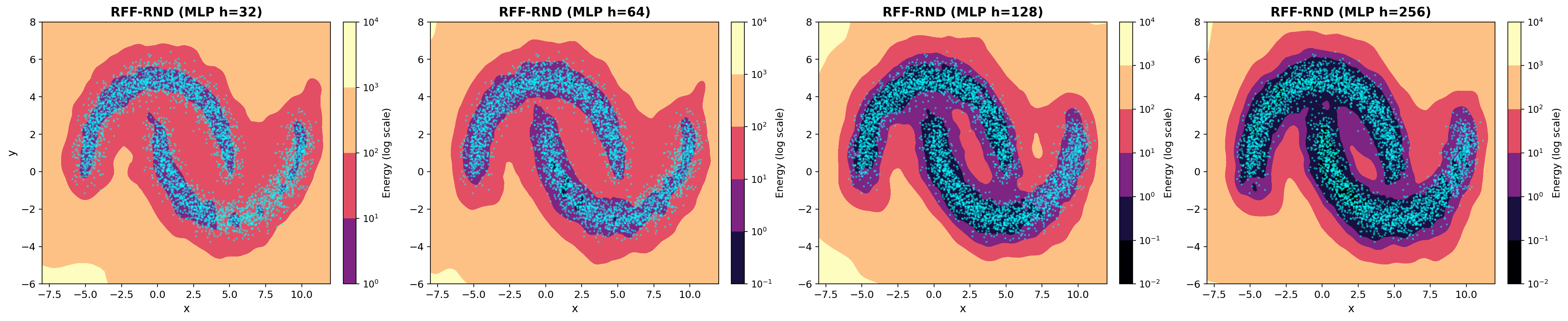}
    \caption{
    \textbf{Effect of predictor width on \texttt{RFF-RND} energy landscapes.}
    We vary the MLP hidden width \(h\in\{32,64,128,256\}\) while keeping the RFF
    target fixed. Across widths, \texttt{RFF-RND} produces a localized energy
    landscape that separates the data manifold from off-manifold regions. Wider
    predictors sharpen the low-energy region around the sampled manifold, but the
    overall manifold-aware structure remains consistent.
    }
    \label{fig:rff_mlp_size_comparison}
\end{figure}

\paragraph{Effect of predictor width \(h\).}
Figure~\ref{fig:rff_mlp_size_comparison} varies the predictor hidden width while keeping the RFF target fixed. Across widths, \texttt{RFF-RND} preserves a
localized energy structure around the data manifold. Increasing \(h\) slightly smooths the energy contours near the sampled manifold, but the overall manifold-aware separation remains stable. This suggests that the qualitative geometry is primarily controlled by the target kernel, while predictor width mainly affects the smoothness of the fitted interpolation.

\clearpage
\newpage
\section{Additional Ablations on RND variants}
\label{app:overconf_MLP_family}

\subsection{RND Variants}
\paragraph{Distributional RND \cite{yang2024exploration}.} We first describe the structure of DRND in Algorithm \ref{alg:drnd}.

\begin{table}[b]
\centering
\caption{DRND Hyperparameters}
\label{tab:drnd_hyperparams}
\begin{tabular}{ll}
\toprule
\textbf{Parameter} & \textbf{Value} \\
\midrule
Input dimension & 2 \\
Feature dimension ($m$) & 32 \\
Number of target networks ($N$) & 10 \\
Hidden dimension & 64 \\
Network depth & 2 \\
Mixing parameter ($\beta$) & 0.9 \\
\midrule
Learning rate & $1\text{e-}4$ \\
Batch size & 128 \\
Training epochs & 5,000 \\
Optimizer & Adam \\
\bottomrule
\end{tabular}
\end{table}

\begin{algorithm}[h]
\caption{Distributional Random Network Distillation (DRND)}
\begin{algorithmic}
    \State \textbf{Input:} Sample $x$; frozen random target networks $\{\tilde{f}_1, \tilde{f}_2, \ldots, \tilde{f}_N\}$; trainable predictor $f_\theta$; step size $\eta$; aggregation weight $\beta \in [0,1]$.
    \State \textbf{Output:} Novelty score $b(x)$ and updated predictor parameters $\theta$.
    
    \State
    \State \textbf{// Training Phase}
    \State Sample target index uniformly: $u(x) \sim \text{Unif}\{\tilde{f}_1(x), \tilde{f}_2(x), \ldots, \tilde{f}_N(x)\}$
    \State Compute prediction: $f_\theta(x)$
    \State Compute MSE loss: $L(\theta) = \|f_\theta(x) - u(x)\|_2^2$
    \State Update predictor: $\theta \gets \theta - \eta \nabla_\theta L(\theta)$
    
    \State
    \State \textbf{// Novelty Computation}
    \State Compute ensemble statistics:
    \State \quad $\mu(x) = \mathbb{E}[X] = \frac{1}{N}\sum_{i=1}^N \tilde{f}_i(x)$ 
    \State \quad $B_2(\mathbf{x}) = \mathbb{E}[X^2] = \frac{1}{N}\sum_{i=1}^N \|\tilde{f}_i(x)\|^2$
    
    \State
    \State Compute normalized deviation:
    \State \quad $b_2(x) = \sqrt{\frac{\|f_\theta(x)\|_2^2 - \|\mu(x)\|_2^2}{B_2(x) - \|\mu(x)\|_2^2}}$
    
    \State
    \State Compute final novelty score:
    \State \quad $b(x) = \beta \|f_\theta(x) - \mu(x)\|^2 + (1-\beta) b_2(x)$
    
    \State \textbf{Return:} $b(x)$
\end{algorithmic}
\label{alg:drnd}
\end{algorithm}

\begin{figure}[t]
    \centering
    \includegraphics[width=1\linewidth]{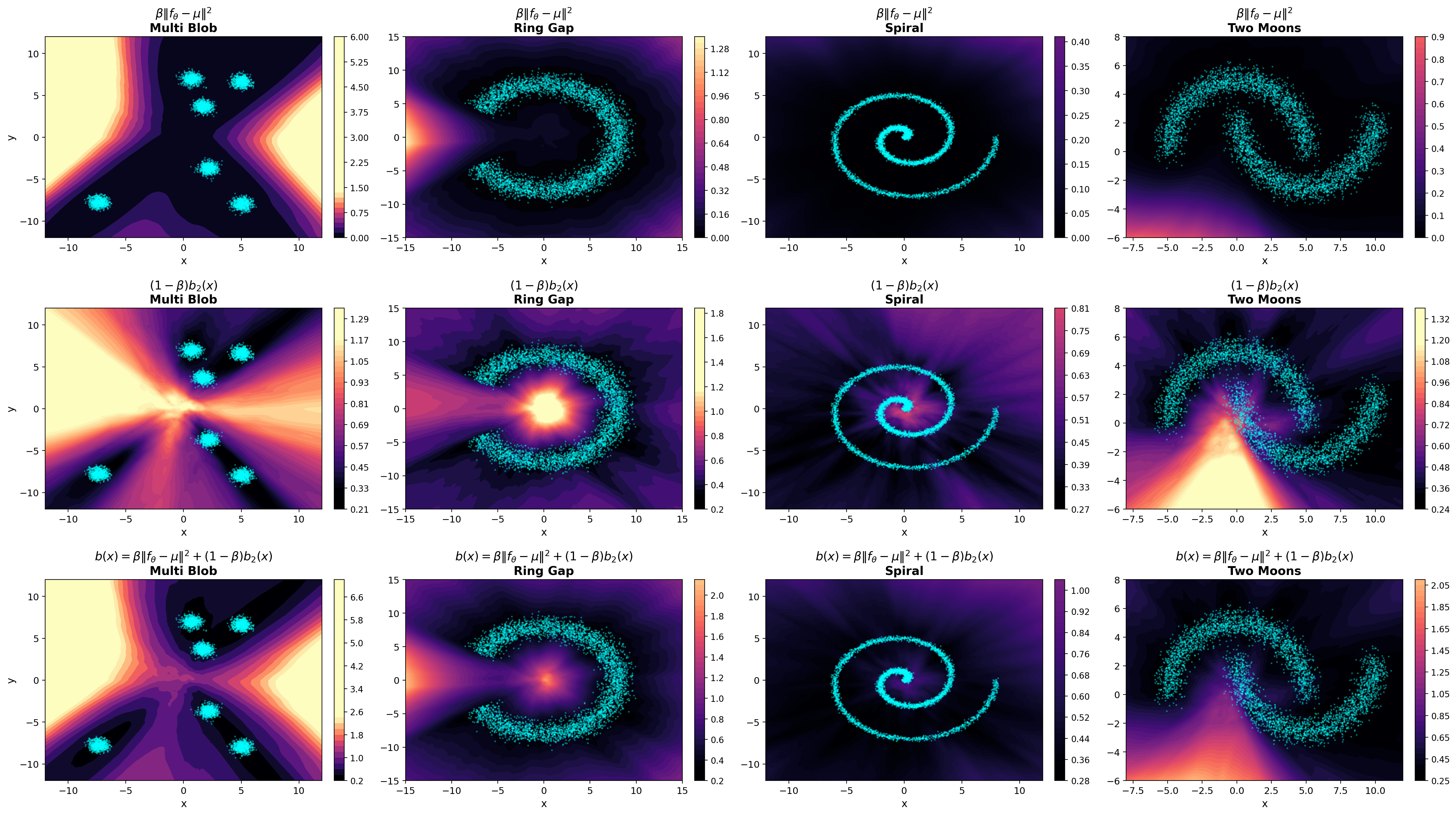}
    \caption{ Energy decomposition of \texttt{DRND} on synthetic manifolds. The first row shows the base RND-like component $\beta \lVert f_\theta(x) - \mu(x) \rVert^2$, which provides a coarse separation between data and off-manifold regions but remains overly smooth. The second row shows the distributional correction term $(1-\beta)b_2(x)$, which partially sharpens the energy landscape by reweighting prediction errors. The third row shows the combined novelty signal $b(x)=\beta \lVert f_\theta(x) - \mu(x) \rVert^2 + (1-\beta)b_2(x)$. While the correction term improves contrast in simple settings such as Multi-Blob, DRND remains insufficiently sensitive to local manifold geometry in more structured datasets, including Ring, Spiral, and Two-Moons.}
    \label{fig:drnd_rdd_comparison_4manifolds}
\end{figure}

The hyperparameters used for DRND are reported in Table \ref{tab:drnd_hyperparams}. Figure~\ref{fig:drnd_rdd_comparison_4manifolds} visualizes the decomposition of the DRND novelty signal into its base and distributional components. Due to a scale mismatch, we do not apply logarithmic scaling to DRND in this visualization. The base term $\beta \lVert f_\theta(x) - \mu(x) \rVert^2$ closely resembles vanilla RND, yielding a smooth energy landscape that separates data-supported regions from off-manifold areas but fails to resolve fine-grained geometric structure. The distributional correction term $(1-\beta)b_2(x)$ partially compensates for this behavior by amplifying certain regions of high prediction uncertainty. This effect is most apparent in the Multi Blob setting, where global density differences align with the correction term and lead to improved separation between clusters.

However, for manifolds with pronounced local geometry, such as Ring Gap, Spiral, and Two Moons, the correction term does not reliably capture interior holes, winding structures, or near-boundary regions. As a result, the combined novelty signal remains dominated by global reweighting rather than locality-aware discrimination. These observations indicate that while DRND improves calibration over vanilla RND, it does not explicitly alter the underlying kernel geometry, motivating approaches that directly control locality through the construction of the induced kernel.

\paragraph{Input-conditioned RND \cite{nikulin2023anti}.}
% In our experiments, the input domain is two-dimensional,
% $x=(x,y)\in\mathbb R^2$. Conditioning is therefore not used to incorporate external side information, but to induce \emph{input-dependent modulation} of the target representation. As a result, each conditioning mechanism defines a different spatially varying transformation over the input space.
There are three main types of neural architectures used in this variant of RND.

\smallskip
\noindent\textbf{Gating.}
Elementwise multiplicative modulation of two learned feature maps:
\[
f(x)
=
\tanh(\mathbf W_1 x + \mathbf b_1)
\;\odot\;
(\mathbf W_2 x + \mathbf b_2).
\]
This form induces axis-aligned, locally smooth variations in the energy
landscape, but remains limited in expressing curved or anisotropic structures.

\smallskip
\noindent\textbf{Bilinear.}
Second-order interaction between input coordinates:
\[
f(x)
=
x^\top \mathbf W x
\;+\;
\mathbf V x
\;+\;
\mathbf b,
\]
where $\mathbf W$ is a third-order tensor.
This allows explicit coupling between $x=(x,y)$ and can induce sharp,
direction-dependent distortions of the novelty signal.

\smallskip
\noindent\textbf{FiLM.}
Input-dependent affine modulation of hidden activations $\mathbf h(x)$:
\[
f(\mathbf h(x))
=
\gamma(x)\odot \mathbf h(x)
\;+\;
\beta(x).
\]
Essentially, FiLM acts as a low-rank bilinear modulation, yielding smoother,
globally consistent deformations of the energy landscape.
\begin{figure}[h]
    \centering
    \includegraphics[width=0.9\linewidth]{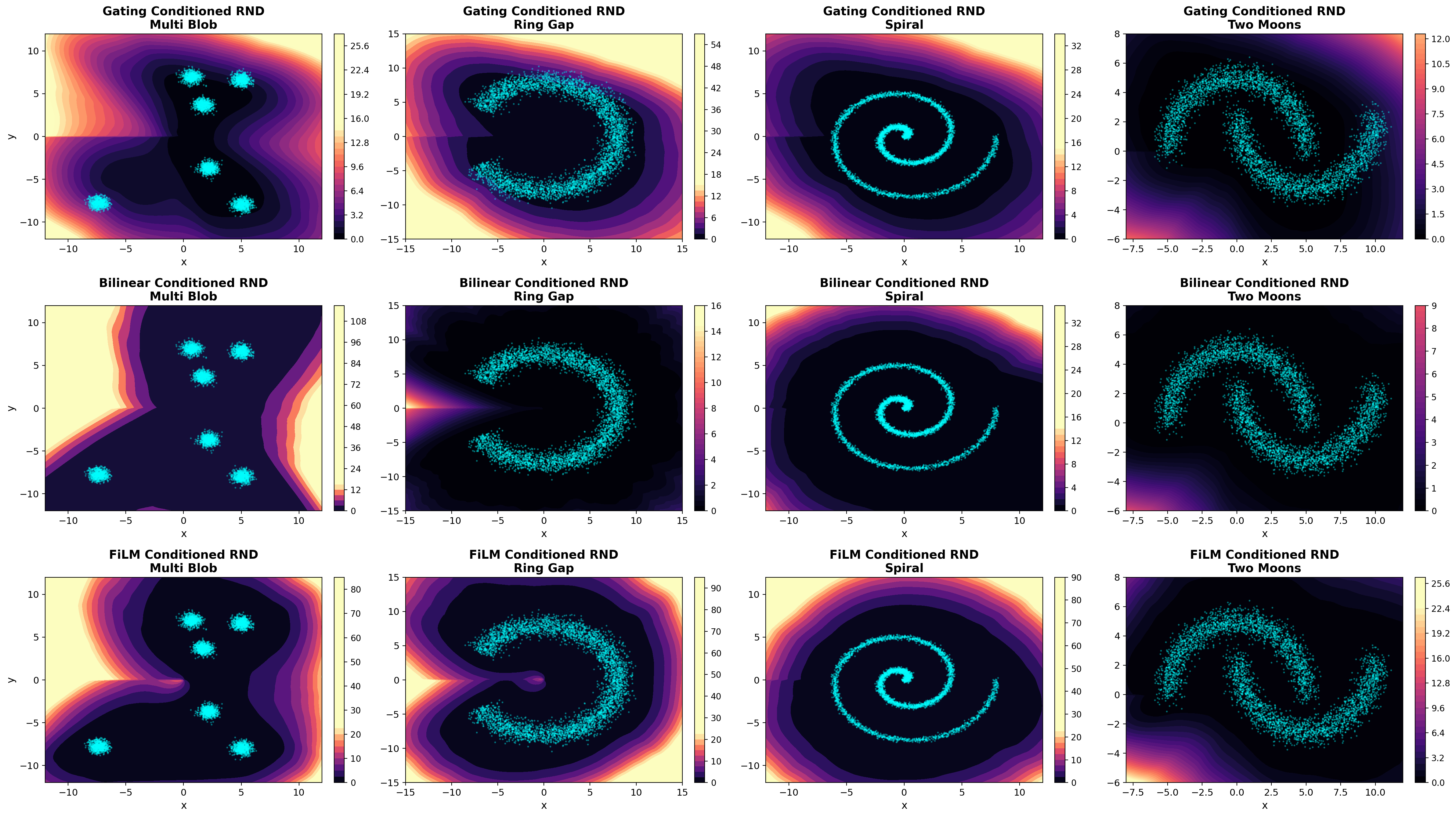}
    \caption{Comparison of input-dependent conditioning mechanisms in RND on two-dimensional synthetic manifolds. Rows correspond to Gating-, Bilinear-, and FiLM-conditioned RND, respectively, while columns show Multi-Blob, Ring Gap, Spiral, and Two-Moons datasets. Background contours visualize the RND energy over the input space, and cyan points denote training samples. Gating induces locally smooth but axis-aligned modulation, Bilinear conditioning yields sharp, direction-dependent distortions, and FiLM produces smoother, globally consistent energy landscapes that better respect curved manifold geometry.}
    \label{fig:conditioned_rnd_comparison_3methods}
\end{figure}

In Figure \ref{fig:conditioned_rnd_comparison_3methods}, we visualize the RND energy over a two-dimensional input space. Since the conditioning signal is derived from the input itself, each method effectively induces a different input-dependent transformation of the space. Gating-conditioned RND produces relatively smooth energy variations, but the modulation remains largely axis-aligned and struggles to adapt to strongly curved manifolds such as spirals. Bilinear conditioning introduces explicit second-order interactions between input coordinates, leading to sharp and highly directional distortions. While expressive, this often results in unstable or overly aggressive deformation of the energy landscape. In contrast, FiLM-conditioned RND yields smoother and more globally coherent energy landscapes. The induced modulation better follows the intrinsic geometry of curved manifolds, preserving low energy regions along the data support while maintaining separation across low-density gaps.

\subsection{Target-Side Control versus Predictor-Side Modulation}
DRND maintains $N$ frozen target MLPs $\{f_i\}_{i=1}^N$ and trains a
single predictor $f_\theta$ to track a uniformly sampled target at each
step; novelty is computed as a deviation statistic $b(x)$ relative to the
target ensemble mean $\mu(x)$.
The target-induced kernel is therefore determined by the MLP function class
at initialization---an NTK-regime quantity that is not analytically
characterized---and evaluating $b(x)$ requires a full forward pass through
all $N$ networks.

Input-conditioned RND variants (gating, bilinear, FiLM) modulate the
\emph{predictor}'s function space via a conditioning signal while keeping
the target network unconditional; the resulting novelty geometry is shaped
jointly by the modulation architecture and the choice of conditioning
representation (e.g., hand-crafted polar coordinates), neither of which
admits a closed-form kernel characterization.

In contrast, \texttt{RFF-RND} replaces the random target network with two
fixed buffers $\Omega\!\in\!\mathbb{R}^{d\times m}$ and $\mathbf{b}\!\in\!\mathbb{R}^m$,
producing features $\varphi(x)=\sqrt{2}\cos(\Omega^\top x+\mathbf{b})$ that are a
Bochner-consistent approximation of a Gaussian kernel with bandwidth
$\alpha$~\citep{rahimi2007random}.
The target-side geometry is therefore explicit and analytically specified:
adjusting $\alpha$ directly controls the length scale of the induced
kernel, without additional ensemble members or modulation modules.

\newpage

\subsection{RND Variants: Architecture and Cost}
% \label{app:overconf_MLP_family}

Table~\ref{tab:arch_summary} summarizes the target and predictor
architecture of each method.
RND and all input-conditioned variants (Gating, Bilinear, FiLM) maintain a
single frozen target MLP, whereas DRND stores $N$ frozen target networks
$\{\tilde{f}_i\}_{i=1}^N$.
\texttt{RFF-RND} replaces the target network entirely with two fixed buffers
$\Omega \in \mathbb{R}^{d \times m}$ and $\mathbf{b} \in \mathbb{R}^m$, with no
gradient or optimizer state.

Table~\ref{tab:cost_summary} reports parameter counts and per-sample
inference complexity under the default configuration ($d{=}2$, $h{=}64$,
$m{=}64$, $N{=}10$, $L{=}2$).
DRND incurs the largest memory footprint (70,752 parameters), with
target-side inference scaling as $\mathcal{O}(NLh^2)$ due to the full
forward pass required through all $N$ networks.
Among the input-conditioned variants, Bilinear is the most expensive
predictor-side at $\mathcal{O}(d^2 m)$ from the third-order tensor
contraction, while Gating and FiLM remain at $\mathcal{O}(hm)$.
\texttt{RFF-RND} matches standard RND in total parameter count (8,704 vs.\
17,024) while reducing target-side cost to $\mathcal{O}(dm)$---a single
matrix multiply followed by an elementwise cosine.

\begin{table}[h]
\centering
\caption{Architecture summary of compared methods.
  $d$: input dim, $h$: hidden dim, $m$: feature dim,
  $N$: number of targets (DRND only), $L$: number of hidden layers.}
\label{tab:arch_summary}
\small
\begin{tabular}{lll}
\toprule
\textbf{Method} & \textbf{Target (frozen)} & \textbf{Predictor} \\
\midrule
RND
  & MLP$(d \to h^L \to m)$
  & MLP$(d \to h^L \to m)$ \\
DRND
  & $N$ frozen MLPs $\{\tilde{f}_i\}_{i=1}^{N}$
  & MLP$(d \to h^L \to m)$ \\
Gating
  & MLP$(d \to h^L \to m)$
  & $\tanh(\mathbf{W}_1 x + \mathbf{b}_1) \odot (\mathbf{W}_2 x + \mathbf{b}_2)$ \\
Bilinear
  & MLP$(d \to h^L \to m)$
  & $x^\top \mathbf{W} x + \mathbf{V}x + \mathbf{b}$,\; $\mathbf{W}\in\mathbb{R}^{d\times d\times m}$ \\
FiLM
  & MLP$(d \to h^L \to m)$
  & $\gamma(x)\odot h(x) + \beta(x)$ \\
\texttt{RFF-RND}
  & $\varphi(x)=\sqrt{2}\cos(\Omega^\top x+\mathbf{b})$,\;
    $\Omega\!\in\!\mathbb{R}^{d\times m}$, $\mathbf{b}\!\in\!\mathbb{R}^m$
  & MLP$(d \to h^L \to m)$ \\
\bottomrule
\end{tabular}
\end{table}

\begin{table}[h]
\centering
\caption{Parameter counts and per-sample inference complexity.
  Defaults: $d{=}2$, $h{=}64$, $m{=}64$, $N{=}10$, $L{=}2$.}
\label{tab:cost_summary}
\small
\begin{tabular}{lrrrll}
\toprule
\textbf{Method}
  & \textbf{Trainable}
  & \textbf{Frozen}
  & \textbf{Total}
  & \textbf{Target} $\mathcal{O}(\cdot)$
  & \textbf{Predictor} $\mathcal{O}(\cdot)$ \\
\midrule
RND
  & 8,512 & 8,512 & 17,024
  & $\mathcal{O}(Lh^2)$
  & $\mathcal{O}(Lh^2)$ \\
DRND
  & 6,432 & 64,320 & 70,752
  & $\mathcal{O}(NLh^2)$
  & $\mathcal{O}(Lh^2)$ \\
Gating
  & 2,368 & 6,432 & 8,800
  & $\mathcal{O}(Lh^2)$
  & $\mathcal{O}(hm)$ \\
Bilinear
  & 6,432 & 6,432 & 12,864
  & $\mathcal{O}(Lh^2)$
  & $\mathcal{O}(d^2 m)$ \\
FiLM
  & 6,816 & 6,432 & 13,248
  & $\mathcal{O}(Lh^2)$
  & $\mathcal{O}(hm)$ \\
\texttt{RFF-RND}
  & 8,512 & $192^\dagger$ & 8,704
  & $\mathcal{O}(dm)$
  & $\mathcal{O}(Lh^2)$ \\
\bottomrule
\end{tabular}

\smallskip
\raggedright\footnotesize
$^\dagger$ Fixed buffers $(\Omega, b)$; no gradient, no optimizer state.\\
DRND target cost scales as $\mathcal{O}(NLh^2)$; all others maintain a single frozen target.
Bilinear predictor incurs $\mathcal{O}(d^2 m)$ from the third-order tensor contraction $x^\top \mathbf{W} x$.
\texttt{RFF-RND} target reduces to a single matmul and elementwise cosine, $\mathcal{O}(dm)$.
\end{table}

\newpage

\section{Detailed Hyperparameters for RL Experiments}
\label{sec:hyperparameters}
For all RL experiments, we use the log-scaled RND signal
\[
b^{+}(\cdot) = \log\!\bigl(1+\mathcal E(\cdot)\bigr)
\]
as the intrinsic reward in exploration settings and as the intrinsic penalty
signal in anti-exploration settings. The constant \(1\) ensures numerical
stability and keeps the logarithm well-defined for nonnegative energies. We
adopt this scaling because raw RND energies can vary substantially across
domains and input modalities, making the corresponding reward or penalty
coefficient difficult to tune. Log-scaling is a common practical remedy in
RND-style methods and is applied uniformly across all RND variants in our
experiments.

\subsection{Online RL Setting}
\paragraph{Proximal Policy Optimization (PPO).}
PPO is an on-policy actor--critic method that alternates between collecting rollouts with a stochastic policy (actor) and improving both the policy and a state-value function (critic) using minibatch stochastic gradient descent. Let the actor be a parameterized policy $\pi_{\theta_{\mathrm{actor}}}(a\mid s)$ and the critic be a value function $V_{\phi_{\mathrm{critic}}}(s)$. Given trajectories $\{(s_t,a_t,r_t)\}_{t=0}^{T-1}$ sampled from the current policy $\pi_{\theta_{\mathrm{actor}}}^-$, PPO forms advantage estimates $\hat A_t$ (e.g., via GAE) using temporal-difference residuals $\delta_t=r_t+\gamma V_{\phi_{\mathrm{critic}}}(s_{t+1})-V_{\phi_{\mathrm{critic}}}(s_t)$. The actor is updated by maximizing the clipped surrogate objective
\begin{equation}
\label{eq:ppo_actor}
\mathcal L_{\mathrm{PPO}}(\theta_{\mathrm{actor}})
\;=\;
\mathbb E_t\!\left[
\min\!\Bigl(
\rho_t(\theta_{\mathrm{actor}})\hat A_t,\;
\mathrm{clip}(\rho_t(\theta_{\mathrm{actor}}),1-\epsilon_{\mathrm{clip}},1+\epsilon_{\mathrm{clip}})\hat A_t
\Bigr)
\right]
+\beta_{\mathrm{ent}}\,\mathbb E_t\!\left[\mathcal H(\pi_{\theta_{\mathrm{actor}}}(\cdot\mid s_t))\right],
\end{equation}
where $\rho_t(\theta_{\mathrm{actor}})=\pi_{\theta_{\mathrm{actor}}}(a_t\mid s_t)/\pi_{\theta_{\mathrm{actor}}^-}(a_t\mid s_t)$, $\epsilon_{\mathrm{clip}}$ is the clip parameter, and the entropy bonus (weight $\beta_{\mathrm{ent}}$) promotes exploration. The critic is trained by regression to a bootstrap target
% \begin{equation}
% \label{eq:ppo_critic}
% \min_{{\phi_{\mathrm{critic}}}}\;
% \mathbb E_t\!\left[\bigl(V_{\phi_{\mathrm{critic}}}(s_t)-\hat V_t\bigr)^2\right],
% \qquad
% \hat V_t \;\approx\; \sum_{k\ge 0}\gamma^k r_{t+k}
% \ \text{(or a GAE-based return)}.
% \end{equation}

\begin{equation}
\label{eq:ppo_critic}
\min_{\phi_{\mathrm{critic}}}\;
\mathbb E_t
\left[
\bigl(
V_{\phi_{\mathrm{critic}}}(s_t)
-
\hat V_t
\bigr)^2
\right],
\qquad
\hat V_t
=
\hat A_t
+
V_{\phi_{\mathrm{critic}}^-}(s_t),
\end{equation}

In practice, we collect a batch of on-policy rollouts, compute $\hat A_t$ and $\hat V_t$, and perform multiple epochs of minibatch updates of \eqref{eq:ppo_actor}--\eqref{eq:ppo_critic}.
\paragraph{Online RL hyperparameters.}
For online RL, Table~\ref{tab:online_hparams} reports the common PPO and
training hyperparameters used across the Atari and Fetch Robotics experiments.
These include learning rates, discount factor, GAE coefficient, rollout length,
clip coefficient, number of environments, network architecture, and the intrinsic
reward coefficient. Task-specific \texttt{RFF-RND} hyperparameters for Atari and Fetch Robotics, including the output dimension \(m\) and bandwidth \(\alpha\), are discussed in Appendix~\ref{sec:additionalExperimentalResults}. 
% In particular,
% Appendix~\ref{subsec:alpha_lambda_performance} reports the hyperparameter search
% protocol and the selected values used in the online experiments.

% \begin{table}[h]
% \centering
% \caption{Hyperparameters for online Atari experiments and Fetch Robotics experiments.(Section~\ref{subsec:online_RL}).}
% \label{tab:online_m_alpha}
% \setlength{\tabcolsep}{12pt}
% \renewcommand{\arraystretch}{1.1}
% \begin{tabular}{l|cc}
% \hline
% \textbf{Environment} & Output dimensions ($m$)  & $\alpha$ \\
% \hline
% Atari MontezumaRevenge & 64 & 1 \\
% Atari Venture          & 64 & 2 \\
% Atari Frogger          & 64 & 2 \\
% Atari Freeway          & 64 & 1 \\
% Fetch Reach            & 64 & 1 \\
% Fetch Push             & 64 & 1 \\
% Fetch Slide            & 64 & 1 \\
% Fetch Pick and Place   & 64 & 1 \\
% \hline
% \end{tabular}
% \end{table}

\begin{table}[h]
\centering
\caption{Hyperparameters of online experiments (Section~\ref{subsec:online_RL})}
\label{tab:online_hparams}
\setlength{\tabcolsep}{8pt}
\renewcommand{\arraystretch}{1.1}
\begin{tabular}{l|l}
\hline
\textbf{Parameters} & \textbf{Value} \\
\hline
Learning rate of the actor network        & $1\text{e-}4$ \\
Learning rate of the critic network       & $1\text{e-}4$ \\
Learning rate of the \texttt{RFF-RND} network         & $3\text{e-}4$  \\
Optimizer                                 & \texttt{Adam} \\
Discount factor ($\gamma$)                & 0.99 \\
GAE coefficient ($\lambda_{\mathrm{GAE}}$)                          & 0.95 \\
Rollout length ($T$)                            & 256 \\
clip coefficient ($\epsilon_{\mathrm{clip}}$)                    & 0.1 \\
Number of environments                   & 32 \\
Actor--Critic network ($\pi_{\theta_{\mathrm{actor}}}$, $V_{\phi_{\mathrm{critic}}}$) & 3-layer CNN encoder + shared MLP \\
Activation function                      & \texttt{ReLU} \\
Number of training epochs                & 4 \\
Intrinsic reward coefficient ($\lambda_{\mathrm{coeff}}$)            & 1 \\
% Extrinsic reward coefficient             & 2 \\
\hline
\end{tabular}
\end{table}

\subsection{Offline RL Setting}
\label{subsec:OfflineRLSettinig}
\paragraph{Soft Actor--Critic (SAC).}
SAC is an off-policy actor--critic method that learns a stochastic policy by
maximizing both expected return and entropy, using replayed transitions.
The actor is a stochastic policy $\pi_{\theta_{\mathrm{actor}}}(a\mid s)$, and the critic consists
of one or more state--action value functions $Q_{\phi}(s,a)$ (often two critics
to mitigate overestimation), together with slowly-updated target networks
$\bar\phi$.
Given transitions $(s,a,r,s')$ sampled from a replay buffer, the critic is
trained to satisfy the soft Bellman equation
\begin{equation}
\label{eq:sac_critic}
\min_{\phi}\;
\mathbb E\!\left[
\bigl(Q_\phi(s,a)-y(s,a,r,s')\bigr)^2
\right],
\quad
y \;=\;
r+\gamma\,\mathbb E_{a'\sim\pi_{\theta_{\mathrm{actor}}}(\cdot\mid s')}\!\left[
\min_i Q_{\bar\phi_i}(s',a') - \alpha_{\mathrm{ent}}\log\pi_{\theta_{\mathrm{actor}}}(a'\mid s')
\right],
\end{equation}
where $\alpha_{\mathrm{ent}}>0$ is the temperature controlling the entropy term.
The actor is updated to maximize the soft objective, equivalently minimizing
\begin{equation}
\label{eq:sac_actor}
\min_{\theta_{\mathrm{actor}}}\;
\mathbb E_{s\sim\mathcal B,\,a\sim\pi_{\theta_{\mathrm{actor}}}(\cdot\mid s)}\!\left[
\alpha_{\mathrm{ent}}\log\pi_{\theta_{\mathrm{actor}}}(a\mid s) - \min_i Q_{\phi_i}(s,a)
\right].
\end{equation}
Optionally, $\alpha_{\mathrm{ent}}$ is tuned by minimizing a dual objective to
match a target entropy. Training alternates between sampling minibatches from
the replay buffer, updating critics via \eqref{eq:sac_critic}, updating the actor
via \eqref{eq:sac_actor}, and updating target networks with Polyak averaging.

\paragraph{Offline RL hyperparameters.}
For offline RL, Table~\ref{tab:offline_hparams} reports the common optimizer,
architecture, and training hyperparameters used across the D4RL experiments.
These include learning rates, discount factors, batch size, network dimensions,
and the number of training updates. Dataset-specific \texttt{RFF-RND} coefficients are discussed separately in
Appendix~\ref{sec:additionalExperimentalResults}. In particular, that section
reports the selected values of
\(\lambda_{\mathrm{coeff}}\) and \(\alpha\) for D4RL locomotion tasks, as well as
\(\lambda_a\), \(\lambda_c\), and \(\alpha\) for AntMaze tasks.

% \begin{table}[h]
% \centering
% \caption{Hyperparameters of D4RL offline experiments}
% \label{tab:offline_hparams}
% \setlength{\tabcolsep}{8pt}
% \renewcommand{\arraystretch}{1.1}
% \begin{tabular}{l|l|l}
% \hline
% \textbf{Name} & \textbf{Parameters} & \textbf{Value} \\
% \hline
% $\eta_{\text{actor}}$  & learning rate of the actor network & $1\text{e-}3$  \\
% $\eta_{\text{critic}}$ & learning rate of the critic network & $1\text{e-}3$ \\
% $\eta_{\text{drnd}}$   & learning rate of the GFF-RND network & $1\text{e-}6$\\
% optimizer             & type of optimizer & Adam \\
% target entropy        & target entropy of the actor & $-\text{action\_dim}$ \\
% $\tau$                & soft update rate & 0.005 \\
% $\gamma$              & discount return & 0.99  \\
% $bs$                  & batch size of the dataset & 1024 \\
% $h$                   & number of hidden layer dimensions & 256 \\
% $e$                   & number of \texttt{RFF-RND} output dimensions & 32 \\
% $n$                   & number of hidden layers & 4 \\
% $f$                   & activation function & ReLU \\
% $K$                   & number of \texttt{RFF-RND} training epochs & 100 \\
% $M$                   & maximum iteration number of SAC & 3000 \\
% $I$                   & gradient updates per iteration & 1000 \\

% \hline
% \end{tabular}
% \end{table}

\begin{table}[h]
\centering
\caption{Hyperparameters of D4RL offline experiments (Section~\ref{subsec:offline_RL}).}
\label{tab:offline_hparams}
\setlength{\tabcolsep}{8pt}
\renewcommand{\arraystretch}{1.1}
\begin{tabular}{l|l}
\hline
\textbf{Parameters} & \textbf{Value} \\
\hline
Learning rate of the actor network & $1\text{e-}3$ ($1\text{e-}4$ on AntMaze) \\
Learning rate of the critic network & $1\text{e-}3$ ($1\text{e-}4$ on AntMaze) \\
Learning rate of the \texttt{RFF-RND} network & $1\text{e-}6$ ($1\text{e-}5$ on AntMaze) \\
 Optimizer & \texttt{Adam} \\
 Activation function & \texttt{ReLU} \\
 Target entropy of the actor & $\texttt{-action\_dim}$ \\
 Soft update rate & 0.005 \\
 Discount factor ($\gamma$) & 0.99 (0.999 on AntMaze) \\
 Batch size & 1,024 \\
 Actor, Critic Hidden dimension & 256 \\
 Actor, Critic Network depth & 3 \\
 \texttt{RFF-RND} hidden dimension & 256 \\
 \texttt{RFF-RND} predictor network depth & 4 \\
 \texttt{RFF-RND} output dimensions ($m$) & 128 \\ %리뷰보고 수정했습니다.
 \texttt{RFF-RND} training epochs & 100 \\
 Maximum iteration number of SAC & 3,000 \\
 Gradient updates per iteration & 1,000 \\
\hline
\end{tabular}
\end{table}

\clearpage
\newpage

\section{Ablation on \texorpdfstring{\texttt{RFF-RND}}{RFF-RND} Hyperparameter Tuning for RL}
\label{sec:additionalExperimentalResults}

\subsection{Concentration Effect of \texorpdfstring{$m$}{m}}
\label{subsec:effect_of_m}
% We compare the performance across different output dimensions $m \in \{32,64,128\}$ on offline D4RL medium datasets. We focus on medium datasets because they provide a balanced setting where both data coverage and policy suboptimality exist, making the effect of representation quality more observable. As shown in Figure~\ref{fig:rnd_energy_heatmap_alpha_vs_dim}, increasing $m$ leads to a smoother kernel energy landscape. This smoother geometry yields more stable novelty estimates, which is particularly beneficial in offline RL where errors cannot be corrected through additional environment interaction.

% Consistent with this observation, Table~\ref{tab:output_dim} shows that $m=128$ achieves the best performance across all three medium datasets. The gains are modest on halfcheetah-medium and hopper-medium, but the improvement is more pronounced on walker2d-medium. This suggests that a higher-dimensional random feature representation provides a more reliable kernel approximation and improves the stability of uncertainty estimation. Overall, increasing $m$ strengthens the robustness of the intrinsic reward signal, although its effect is smaller than the influence of $\alpha$ and $\lambda_{\rm{coeff}}$.

We first study the effect of the RFF output dimension \(m\) on downstream offline RL performance. As discussed in
Section~\ref{subsec:diagnostic_and_ablations_for_rffrnd}, \(m\) controls the
finite-output stability of the RND energy. In particular, increasing \(m\)
reduces finite-output variance and makes the empirical energy
\(\mathcal E_m(x)\) more closely match the kernel-level prediction. The
energy-landscape diagnostics also show that larger \(m\) produces smoother and
more stable empirical novelty estimates, without changing the underlying target
kernel itself.

To verify whether this diagnostic observation translates to downstream
performance, we compare \(m \in \{32,64,128\}\) on offline D4RL medium datasets.
We focus on medium datasets because they provide a balanced setting where both
data coverage and policy suboptimality exist, making the effect of
representation stability more observable. As shown in
Table~\ref{tab:output_dim}, the performance variation across \(m\) is relatively
small, but \(m=128\) achieves the best average performance across the three
offline medium datasets. This suggests that increasing \(m\) improves the
stability of the random feature approximation, although the downstream gain is
marginal once \(m\) is sufficiently large. 

% As shown in Table~\ref{tab:output_dim},
% \(m=128\) achieves the best performance across all three medium datasets. The
% gains are modest on halfcheetah-medium and hopper-medium, but the improvement is
% more pronounced on walker2d-medium. Based on both the diagnostic analysis in
% Section~\ref{subsec:diagnostic_and_ablations_for_rffrnd} and the downstream
% ablation in this section, we fix \(m=128\) for the offline main experiments.

\begin{table}[h]
\centering
\caption{Performance comparison across different output dimensions \(m\) on offline D4RL medium datasets.}
\label{tab:output_dim}
\begin{tabular}{lccc}
\toprule
Dataset & $m=32$ & $m=64$ & $m=128$ \\
\midrule
halfcheetah-medium & $66.3 \pm 1.1$ & $66.9 \pm 2.3$ & \textbf{$67.0 \pm 0.5$} \\
hopper-medium      & $101.5 \pm 2.3$ & $99.5 \pm 6.7$ & \textbf{$102.3 \pm 3.5$} \\
walker2d-medium    & $92.3 \pm 4.1$ & $91.5 \pm 4.6$ & \textbf{$93.7 \pm 2.8$} \\
\midrule
Average            & $86.7$ & $86.0$ & \textbf{$87.7$} \\
\bottomrule
\end{tabular}
\end{table}

We also evaluate the effect of \(m\) in online RL using Atari Freeway and
Frogger. For this validation, each run is trained for 1M environment steps. As
shown in Table~\ref{tab:online_output_dim}, the performance variation across
\(m\) is relatively small once \(m\) is moderately large, and \(m=64\) achieves
the best average performance.

\begin{table}[h]
\centering
\caption{Performance comparison across different output dimensions \(m\) on online Atari tasks trained for 1M update steps.}
\label{tab:online_output_dim}
\begin{tabular}{lccc}
\toprule
Environment & $m=32$ & $m=64$ & $m=128$ \\
\midrule
Freeway & $20.3 \pm 2.5$ & \textbf{$22.8 \pm 1.9$} & $21.5 \pm 2.3$ \\
Frogger & $16.4 \pm 3.7$ & \textbf{$17.1 \pm 2.5$} & $16.8 \pm 3.1$ \\
\midrule
Average & $18.4$ & \textbf{$20.0$} & $19.2$ \\
\bottomrule
\end{tabular}
\end{table}

Overall, changing \(m\) has a marginal effect on downstream performance compared
with the effects of \(\alpha\) and \(\lambda_{\mathrm{coeff}}\). Based on these
validation results, we use \(m=128\) for the offline RL experiments and \(m=64\)
for the online RL experiments.

%6.1절에서의 분석+이subsection에서의 실험으로 m을 고정해두고 시작하겠다.
\subsection{
% \texorpdfstring
Performance Across \texorpdfstring{\(\lambda_{\mathrm{coeff}}\)}{lambda-coeff} and \texorpdfstring{\(\alpha\)}{a}}
% {Performance Across lambda_coeff and alpha}}
\label{subsec:alpha_lambda_performance}

\paragraph{Hyperparameter search protocol.}
We describe the hyperparameter search protocol used for the RL experiments.
Following common practice in prior RND-based offline RL studies
~\citep{nikulin2023anti,tarasov2023revisiting,yang2024exploration}, we use a
grid search to select task-specific hyperparameters.

For the online RL Atari tasks, we fix \(\lambda_{\mathrm{coeff}} = 1\) across
all Atari and Fetch experiments to isolate the effect of the intrinsic reward
geometry. We tune \(\alpha \in \{0.5, 1, 2, 4\}\). The final online
configuration, including the selected environment-specific values, is reported
in Table~\ref{tab:online_m_alpha}.

For the offline D4RL locomotion tasks, we tune
\(\lambda_{\mathrm{coeff}} \in \{1, 2, 3, 5, 10\}\) and
\(\alpha \in \{1, 2, 3\}\). The selected dataset-specific values are reported
in Table~\ref{tab:d4rl_anti_exploration_blank}. For AntMaze, we decouple the actor and critic coefficients and tune
\(\lambda_a \in \{0.001, 0.005, 0.01, 0.05, 0.1, 0.5, 1.0, 2.0\}\),
\(\lambda_c \in \{0.0002, 0.001, 0.005, 0.01, 0.05\}\), and
\(\alpha \in \{0.001, 0.01, 0.02, 0.1\}\). The selected values for AntMaze are
reported in Table~\ref{tab:antmaze_hyperparameters}.

% For online RL Atari tasks, we fix \(\lambda_{\mathrm{coeff}}=1\) across all Atari and Fetch
% experiments to isolate the effect of the intrinsic reward geometry. We tune
% \(\alpha\in\{0.5,1,2,4\}\).
% % and vary \(m\in\{32,64,128\}\). 
% The final online configuration
% % uses \(m=64\), 
% with the selected environment-specific values is reported in Table~\ref{tab:online_m_alpha}. 

% For offline D4RL locomotion tasks,
% we tune \(\lambda_{\mathrm{coeff}}\in\{1,2,3,5,10\}\) and
% \(\alpha\in\{1,2,3\}\).
% % , while fixing \(m=128\).
% The selected dataset-specific
% values are reported in Table~\ref{tab:d4rl_anti_exploration_blank}. 

% For the AntMaze case, we decouple the actor and critic coefficients and tune
% \(\lambda_a \in \{0.001, 0.005, 0.01, 0.05, 0.1, 0.5, 1.0, 2.0\}\),
% \(\lambda_c \in \{0.0002, 0.001, 0.005, 0.01, 0.05\}\), and
% \(\alpha \in \{0.001, 0.01, 0.02, 0.1\}\).
% The selected values for AntMaze are reported in
% Table~\ref{tab:antmaze_hyperparameters}.

\begin{table}[!htbp]
\centering
\caption{Hyperparameters for online Atari and Fetch Robotics experiments (Section~\ref{subsec:online_RL}).}
\label{tab:online_m_alpha}
\setlength{\tabcolsep}{12pt}
\renewcommand{\arraystretch}{1.1}
\begin{tabular}{l|cc}
\hline
\textbf{Environment} &  $\alpha$ \\
\hline
Atari MontezumaRevenge  & 1 \\
Atari Venture           & 2 \\
Atari Frogger           & 2 \\
Atari Freeway           & 1 \\
Fetch Reach             & 1 \\
Fetch Push              & 1 \\
Fetch Slide             & 1 \\
Fetch Pick and Place    & 1 \\
\hline
\end{tabular}
\end{table}

\begin{table}[!htbp]
\centering
\caption{Hyperparameters for D4RL Locomotion offline datasets (Section~\ref{subsec:offline_RL}).}
\label{tab:d4rl_anti_exploration_blank}
\setlength{\tabcolsep}{12pt}
\renewcommand{\arraystretch}{1.1}
\begin{tabular}{l|cc}
\hline
\textbf{Dataset Name} & $\lambda_{\mathrm{coeff}}$ & $\alpha$ \\
\hline

halfcheetah-random & 5 & 1 \\
halfcheetah-medium & 2 & 3 \\
halfcheetah-expert & 10 & 3 \\
halfcheetah-medium-expert & 10 & 3 \\
halfcheetah-medium-replay & 1 & 2 \\
halfcheetah-full-replay & 1 & 3\\
\hline
hopper-random & 1 &  2\\
hopper-medium & 5 & 3 \\
hopper-expert & 5 & 1 \\
hopper-medium-expert & 1 & 1 \\
hopper-medium-replay &1 & 1 \\
hopper-full-replay & 3 & 3 \\
\hline
walker2d-random & 5 & 3 \\
walker2d-medium & 5 & 3 \\
walker2d-expert & 5 & 2 \\
walker2d-medium-expert & 5 & 2 \\
walker2d-medium-replay & 3 & 3 \\
walker2d-full-replay & 1 & 3 \\

\hline
\end{tabular}
\end{table}

\begin{table}[!htbp]
\centering
\caption{Hyperparameters for AntMaze offline datasets (Section~\ref{subsec:offline_RL}).}
\label{tab:antmaze_hyperparameters}
\setlength{\tabcolsep}{10pt}
\renewcommand{\arraystretch}{1.1}
\begin{tabular}{l|ccc}
\hline
\textbf{Dataset Name} & $\lambda_a$ & $\lambda_c$ & $\alpha$ \\
\hline
antmaze-umaze & 0.005 & 0.005 & 0.02 \\
antmaze-umaze-diverse & 0.01 & 0.001 & 0.1 \\
antmaze-medium-play & 2.0 & 0.05 & 0.01 \\
antmaze-medium-diverse & 0.001 & 0.0002 & 0.001 \\
antmaze-large-play & 0.05 & 0.001 & 0.001 \\
antmaze-large-diverse & 0.005 & 0.001 & 0.001 \\
\hline
\end{tabular}
\end{table}

% \paragraph{Focused sweep over \texorpdfstring{$\alpha$}{alpha} and \texorpdfstring{$\lambda_{\mathrm{coeff}}$}{lambda}.}
% To better understand the interaction between \(\alpha\) and
% \(\lambda_{\mathrm{coeff}}\), we perform a focused sweep on representative D4RL
% medium locomotion datasets: halfcheetah-medium, hopper-medium, and
% walker2d-medium. We use \(m=128\), evaluate
% \(\alpha\in\{1,2,3\}\) and
% \(\lambda_{\mathrm{coeff}}\in\{1,2,3,5\}\), and average each result over 5
% random seeds. Tables~\ref{tab:hc_expected}--\ref{tab:walker_expected} show that
% performance is not monotonic in either parameter. On halfcheetah-medium, the
% best result is obtained at \(\alpha=3\) and
% \(\lambda_{\mathrm{coeff}}=2\), while increasing
% \(\lambda_{\mathrm{coeff}}\) further with the same \(\alpha\) causes a large
% performance drop. In contrast, hopper-medium and walker2d-medium benefit from a
% stronger regularization coefficient and achieve the best performance at
% \(\lambda_{\mathrm{coeff}}=5\). These results suggest that the effective level
% of conservatism depends on the dataset dynamics and that \(\alpha\) and
% \(\lambda_{\mathrm{coeff}}\) should be selected jointly.

\newpage

\paragraph{Role of \texorpdfstring{$\alpha$}{alpha} in exploration and anti-exploration.}

\begin{figure*}[!htbp]
\centering
\begin{tabular}{ccc}
\includegraphics[width=0.3\textwidth]{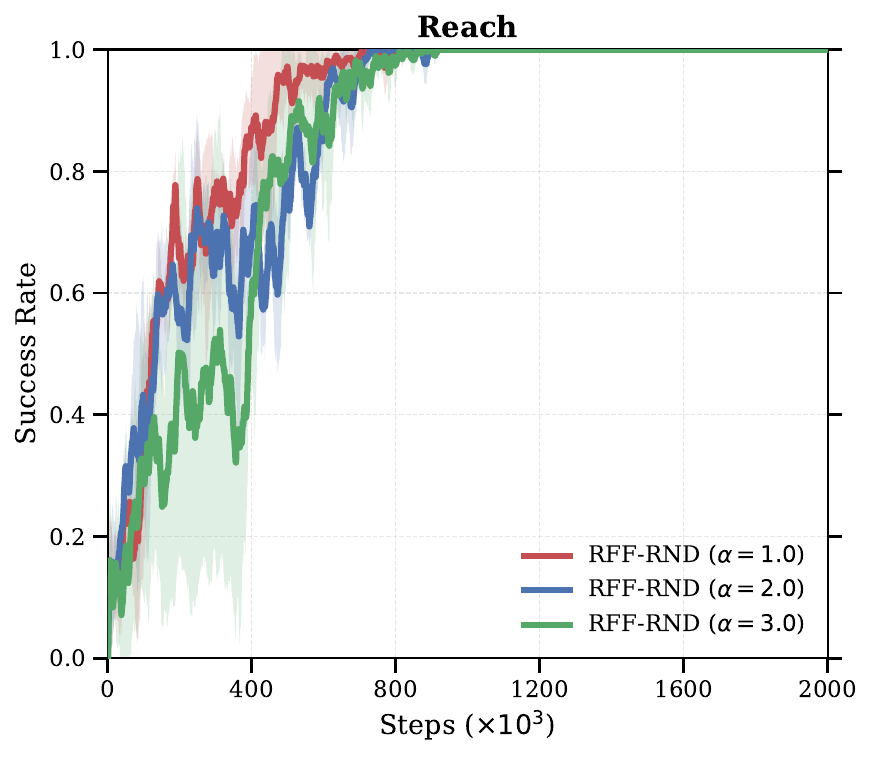} &
\includegraphics[width=0.3\textwidth]{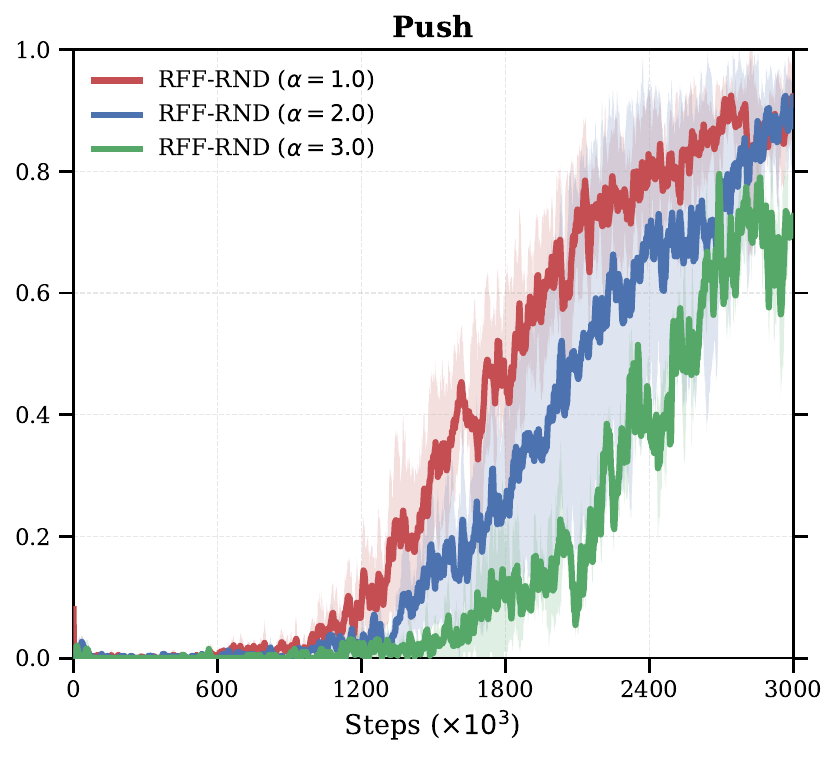} &
\includegraphics[width=0.3\textwidth]{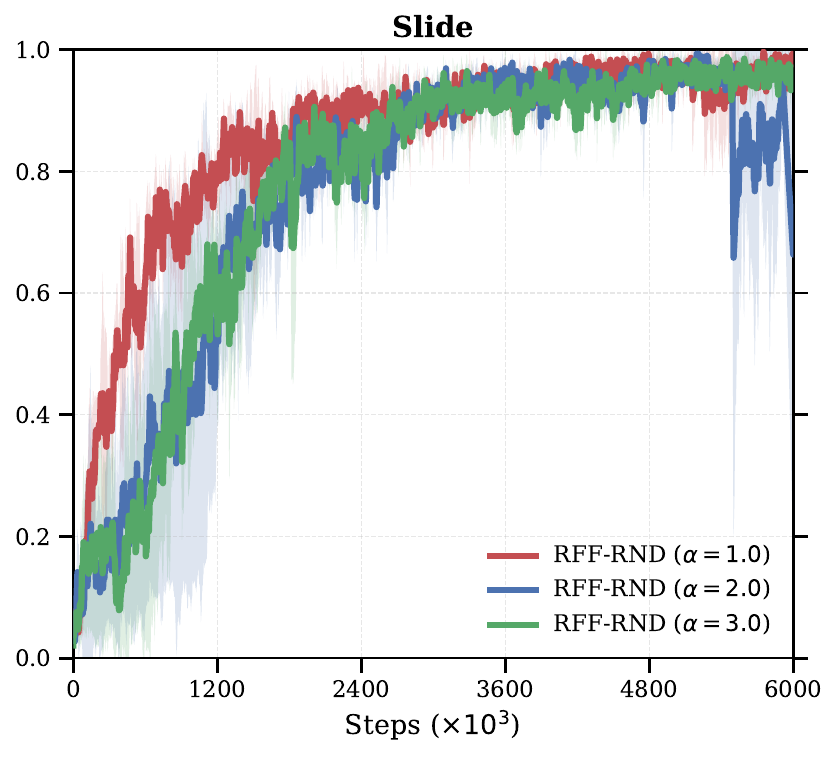} \\
(a) Reach & (b) Push & (c) Slide
\end{tabular}
\caption{Effect of \(\alpha\) on exploration performance in Fetch Robotics tasks.}
\label{fig:alpha_fetch}
\end{figure*}

In online sparse-reward exploration, the preferred behavior can be different.
Smaller \(\alpha\) induces a sharper novelty boundary, which helps the agent
distinguish nearby but unexplored regions more quickly. We validate this on
Fetch Robotics tasks in Figure~\ref{fig:alpha_fetch}. In particular,
\(\alpha=2\) and \(\alpha=3\) reach the 50\% success threshold 76.0\% and
77.5\% later than \(\alpha=1\), respectively. These results show that
\(\alpha\) affects both offline anti-exploration and online exploration, but the
desirable scale depends on the task setting. Overall, the results justify using a
small grid search for \(\alpha\) and \(\lambda_{\mathrm{coeff}}\), followed by
dataset-specific hyperparameter selection.

We also examine the role of \(\alpha\) more directly because it controls the
spatial scale of the kernel-induced novelty estimate. In offline RL, larger
\(\alpha\) produces smoother uncertainty estimates, which can be useful when the
dataset has broad support. Table~\ref{tab:alpha_random} supports this trend on
D4RL random datasets, where \(\alpha=2\) achieves the highest average score.
However, the best value is not identical across all environments, as
halfcheetah-random performs best at \(\alpha=1\). This again indicates that
\(\alpha\) affects anti-exploration behavior but does not follow a universal
monotonic pattern.

\begin{table}[h]
\centering
\caption{Performance comparison across different \(\alpha\) values on D4RL random datasets.}
\label{tab:alpha_random}
\begin{tabular}{lccc}
\toprule
Dataset & $\alpha=0.5$ & $\alpha=1$ & $\alpha=2$ \\
\midrule
halfcheetah-random & 24.6 $\pm$ 1.6 & \textbf{29.3 $\pm$ 1.7} & 29.0 $\pm$ 1.2 \\
hopper-random      & 26.5 $\pm$ 0.2 & 30.8 $\pm$ 0.3 & \textbf{31.4 $\pm$ 0.1} \\
walker2d-random    & 18.9 $\pm$ 0.7 & 17.3 $\pm$ 0.8 & \textbf{21.7 $\pm$ 1.6} \\
\midrule
Average            & 23.3 & 25.8 & \textbf{27.4} \\
\bottomrule
\end{tabular}
\end{table}

\end{document}

%% file: checklist.tex
\section*{NeurIPS Paper Checklist}

%%% BEGIN INSTRUCTIONS %%%
The checklist is designed to encourage best practices for responsible machine learning research, addressing issues of reproducibility, transparency, research ethics, and societal impact. Do not remove the checklist: {\bf The papers not including the checklist will be desk rejected.} The checklist should follow the references and follow the (optional) supplemental material.  The checklist does NOT count towards the page
limit. 

Please read the checklist guidelines carefully for information on how to answer these questions. For each question in the checklist:
\begin{itemize}
    \item You should answer \answerYes{}, \answerNo{}, or \answerNA{}.
    \item \answerNA{} means either that the question is Not Applicable for that particular paper or the relevant information is Not Available.
    \item Please provide a short (1--2 sentence) justification right after your answer (even for \answerNA). 
   % \item {\bf The papers not including the checklist will be desk rejected.}
\end{itemize}

{\bf The checklist answers are an integral part of your paper submission.} They are visible to the reviewers, area chairs, senior area chairs, and ethics reviewers. You will also be asked to include it (after eventual revisions) with the final version of your paper, and its final version will be published with the paper.

The reviewers of your paper will be asked to use the checklist as one of the factors in their evaluation. While \answerYes{} is generally preferable to \answerNo{}, it is perfectly acceptable to answer \answerNo{} provided a proper justification is given (e.g., error bars are not reported because it would be too computationally expensive'' or ``we were unable to find the license for the dataset we used''). In general, answering \answerNo{} or \answerNA{} is not grounds for rejection. While the questions are phrased in a binary way, we acknowledge that the true answer is often more nuanced, so please just use your best judgment and write a justification to elaborate. All supporting evidence can appear either in the main paper or the supplemental material, provided in appendix. If you answer \answerYes{} to a question, in the justification please point to the section(s) where related material for the question can be found.

IMPORTANT, please:
\begin{itemize}
    \item {\bf Delete this instruction block, but keep the section heading ``NeurIPS Paper Checklist"},
    \item  {\bf Keep the checklist subsection headings, questions/answers and guidelines below.}
    \item {\bf Do not modify the questions and only use the provided macros for your answers}.
\end{itemize}

%%% END INSTRUCTIONS %%%

\begin{enumerate}

\item {\bf Claims}
    \item[] Question: Do the main claims made in the abstract and introduction accurately reflect the paper's contributions and scope?
    \item[] Answer: \answerYes{}.
    \item[] Justification: The abstract and introduction state the paper's main contributions: a GP/NTK-based analysis of RND overgeneralization, the resulting spectral interpretation, and the proposed \texttt{RFF-RND} target design. The claims are scoped to RND-style novelty estimation and are supported by the theoretical results and experiments reported in the paper.

\item {\bf Limitations}
    \item[] Question: Does the paper discuss the limitations of the work performed by the authors?
    \item[] Answer: \answerYes{}.
    \item[] Justification: The paper discusses limitations of the present work, including the idealized GP/NTK-regime analysis, the restriction of the theoretical model to target-prediction RND mechanisms, and the fact that the proposed bandwidth-based control is evaluated on a finite set of RL benchmarks. We also discuss practical limitations such as sensitivity to bandwidth selection and the need to further study predictor optimization dynamics and higher-dimensional visual inputs.

    \item {\bf Theory assumptions and proofs}
    \item[] Question: For each theoretical result, does the paper provide the full set of assumptions and a complete (and correct) proof?
    \item[] Answer: \answerYes{}.
    \item[] Justification: The paper states the assumptions for each theoretical result, including the GP target model, predictor interpolation kernel, regularization, and bandwidth-controlled kernel conditions. Complete proofs are provided in the appendix, with the main text giving the corresponding intuition and implications.
    \item {\bf Experimental result reproducibility}
    \item[] Question: Does the paper fully disclose all the information needed to reproduce the main experimental results of the paper to the extent that it affects the main claims and/or conclusions of the paper (regardless of whether the code and data are provided or not)?
    \item[] Answer: \answerYes{}.
    \item[] Justification: The paper describes the experimental tasks, baselines, model classes, evaluation protocol, random seeds, and hyperparameter ranges used for the main experiments. Additional implementation details are provided in the appendix and in the accompanying anonymized code package.


\item {\bf Open access to data and code}
    \item[] Question: Does the paper provide open access to the data and code, with sufficient instructions to faithfully reproduce the main experimental results, as described in supplemental material?
    \item[] Answer: \answerYes{}.
    \item[] Justification: We provide an anonymized code package with scripts and instructions for reproducing the main experimental results. The experiments use standard public RL benchmarks, and the supplemental material describes the required dependencies, data access procedures, and commands for running the proposed method and baselines.


\item {\bf Experimental setting/details}
    \item[] Question: Does the paper specify all the training and test details (e.g., data splits, hyperparameters, how they were chosen, type of optimizer) necessary to understand the results?
    \item[] Answer: \answerYes{}.
    \item[] Justification: The paper specifies the online and offline RL benchmarks, baselines, evaluation metrics, random seeds, training configurations, optimizers, and hyperparameter sweeps used for the proposed method and comparisons. Further details are included in the appendix and accompanying code package.


\item {\bf Experiment statistical significance}
    \item[] Question: Does the paper report error bars suitably and correctly defined or other appropriate information about the statistical significance of the experiments?
    \item[] Answer: \answerYes{}.
    \item[] Justification: The experimental figures and tables report performance over multiple random seeds, with variability shown using standard deviations or shaded regions as specified in the corresponding captions. The reported uncertainty reflects variability across random initialization and environment-level randomness.

\item {\bf Experiments compute resources}
    \item[] Question: For each experiment, does the paper provide sufficient information on the computer resources (type of compute workers, memory, time of execution) needed to reproduce the experiments?
    \item[] Answer: \answerYes{}.
    \item[] Justification: The appendix reports the compute resources used for the experiments, including GPU type, memory, approximate runtime per run, and estimated total GPU hours. We also indicate that the reported compute corresponds to the experiments included in the paper and does not include all preliminary or failed experiments.
    
\item {\bf Code of ethics}
    \item[] Question: Does the research conducted in the paper conform, in every respect, with the NeurIPS Code of Ethics \url{https://neurips.cc/public/EthicsGuidelines}?
    \item[] Answer: \answerYes{}.
    \item[] Justification: The research uses standard reinforcement learning benchmarks and synthetic or publicly available datasets, and does not involve human subjects, private data, or sensitive personal information. We have reviewed the NeurIPS Code of Ethics and believe the work conforms to it.

\item {\bf Broader impacts}
    \item[] Question: Does the paper discuss both potential positive societal impacts and negative societal impacts of the work performed?
    \item[] Answer: \answerYes{}.
    \item[] Justification: The paper discusses the potential positive impact of more reliable novelty estimation for exploration, offline conservatism, and safety-aware decision making. It also notes that improved exploration and autonomy methods should be carefully evaluated before deployment in safety-critical systems such as robotics or autonomous driving.
    
\item {\bf Safeguards}
    \item[] Question: Does the paper describe safeguards that have been put in place for responsible release of data or models that have a high risk for misuse (e.g., pre-trained language models, image generators, or scraped datasets)?
    \item[] Answer: \answerNA{}.
    \item[] Justification: The paper does not release high-risk pretrained models, image generators, language models, or scraped datasets. The experiments use standard RL benchmarks and do not introduce assets with a high risk of misuse.

\item {\bf Licenses for existing assets}
    \item[] Question: Are the creators or original owners of assets (e.g., code, data, models), used in the paper, properly credited and are the license and terms of use explicitly mentioned and properly respected?
    \item[] Answer: \answerYes{}.
    \item[] Justification: The paper cites the original sources for the benchmarks, datasets, and baseline algorithms used in the experiments. The appendix documents the public benchmark environments and software packages used, including their versions and license information where available.

\item {\bf New assets}
    \item[] Question: Are new assets introduced in the paper well documented and is the documentation provided alongside the assets?
    \item[] Answer: \answerYes{}.
    \item[] Justification: The paper introduces an implementation of \texttt{RFF-RND} and provides it as part of the anonymized code package. The accompanying documentation describes the method implementation, dependencies, command-line usage, hyperparameters, and scripts for reproducing the main results.

\item {\bf Crowdsourcing and research with human subjects}
    \item[] Question: For crowdsourcing experiments and research with human subjects, does the paper include the full text of instructions given to participants and screenshots, if applicable, as well as details about compensation (if any)? 
    \item[] Answer: \answerNA{}.
    \item[] Justification: The paper does not involve crowdsourcing, human-subject experiments, user studies, or human annotation.


\item {\bf Institutional review board (IRB) approvals or equivalent for research with human subjects}
    \item[] Question: Does the paper describe potential risks incurred by study participants, whether such risks were disclosed to the subjects, and whether Institutional Review Board (IRB) approvals (or an equivalent approval/review based on the requirements of your country or institution) were obtained?
    \item[] Answer: \answerNA{}.
    \item[] Justification: The paper does not involve human subjects, crowdsourcing, or user studies, so IRB approval or equivalent review is not applicable.

\item {\bf Declaration of LLM usage}
    \item[] Question: Does the paper describe the usage of LLMs if it is an important, original, or non-standard component of the core methods in this research? Note that if the LLM is used only for writing, editing, or formatting purposes and does \emph{not} impact the core methodology, scientific rigor, or originality of the research, declaration is not required.
    \item[] Answer: \answerNA{}.
    \item[] Justification: The core method, analysis, and experiments do not use LLMs as an important, original, or non-standard component. Any use of general writing or editing assistance, if any, does not affect the scientific methodology, rigor, or originality of the work.


\end{enumerate}